**Department of Education and Science, Russian Federation**

**Perm State University**

# FUNDAMENTALS OF MATHEMATICAL THEORY OF EMOTIONAL ROBOTS

**MONOGRAPH**

**Oleg G. Pensky, Kirill V. Chernikov**

2010  Perm, RUSSIA

**Abstract**

In this book we introduce a mathematically formalized concept of emotion, robot's education and other psychological parameters of intelligent robots. We also introduce unitless coefficients characterizing an emotional memory of a robot. Besides, the effect of a robot's memory upon its emotional behavior is studied, and theorems defining fellowship and conflicts in groups of robots are proved. Also unitless parameters describing emotional states of those groups are introduced, and a rule of making alternative (binary) decisions based on emotional selection is given. We introduce a concept of equivalent educational process for robots and a concept of efficiency coefficient of an educational process, and suggest an algorithm of emotional contacts within a group of robots. And generally, we present and describe a model of a virtual reality with emotional robots.

The book is meant for mathematical modeling specialists and emotional robot software developers.

# CONTENTS

# INTRODUCTION

Emotions represent an essential part of human and animal psychological activity.

Attempts to formalize mathematically the psychological behavior of higher living beings were performed in a book «Гипотезы и алгоритмы математической теории исчисления эмоций» ("Hypotheses and Algorithms of Emotion Calculus Mathematical Theory") edited by Professor Oleg G. Pensky and published by the Perm State University (Russia) in 2009. Although the authors wanted this treatise to be considered as an example of some scientific quest, it encountered strong misunderstanding of psychologists in the city of Perm.

That book suggested mathematical models introducing and applying such terms and concepts as 'emotional education/upbringing', 'reeducation', 'temperament', 'conflict', etc.; also the authors reviewed approaches to modeling of emotional behavior of subjects, estimation of a psychological state in groups; there was as well suggested a new approach to the description of some new economic phenomena based on psychological theories.

The authors of the present paper completely agree that computer modeling of emotions is hindered by ambiguity of living being emotional behavior.

Considering misunderstanding of psychologists, Professor Pensky decided to adapt the results of his studies performed in 2009 to mathematical modeling of emotional robots and give a further development to those ideas.

The treatise of professor Oleg G. Pensky titled "Mathematical Models of Emotional Robots" was issued by the Perm State University printing office in 2010.

In the present book, same as in that one issued in 2010, the authors made an attempt to create and mathematically describe a virtual reality of emotional robots, which is based on such key terms as emotions and education, and includes fellowship/concordance and conflicts between its inhabitants–robots which feature various abilities, temperaments, memory, will-power, emotional work under achieving goals, 'diseases', education process prospects and corresponding concepts and terms.

Currently the American scientists [1] work on creation of an electronic copy of a human being which would be called E-creature. By happy chance the present book touches upon those very topics which are currently studied by our American colleagues. We consider robots with a non-absolute memory, and this kind of memory is a human being's feature.

Of course, the mathematical theory of emotional robots which we call your attention to in this book is far from perfection. But its authors never meant that this theory claims to be global, and once again ask critics above all to consider this book as an example of a scientific quest.

*Acknowledgements*

The authors are eternally indebted to Alexander Bolonkin, PhD, Professor of NJIT for having the book discussed, for the description of E-creature information modeling problems, for his guidance in advancing and presenting our theory of emotional robots to the scientific community.

The authors highly appreciate useful notes concerning the content of this book made by Tatiana S. Belozerova, PhD (Russia).

# 1. ROBOT's EMOTION: DEFINITION

A theory of human psychology defines emotions as an organism response to some stimulus [2]. Concerning robots, let us designate this stimulus as '*subject*', and define it as follows:

Let *t* be a time.

<u>Definition 1.1</u>. The function *S(t)* is referred as a 'subject' if it has the following properties:
1. Function domain of *S(t)*: $t \in [0, t^*]$, $t^* > 0$, $t^* < \infty$;
2. *S(t)*>0 for any $t \in [0, t^*]$;
3. *S(t)* is the one-to-one function;
4. *S(t)* is the bounded function.

The paper [3] contains a theorem proving that it is possible for computer software to model human and animal emotions. But psychological features of living beings' emotions are so intricate and ambiguous that we decided to introduce a special mathematical definition of a robot's emotion. In this definition we are abstracting from real human emotions and, at the same time, accumulating general features of human and animal emotions; we are also abstracting from the content of emotions.

<u>Definition 1.2</u>. The function *f(t)*, satisfying the equation $f(t) = a(S(t),t)S(t)$ (with *a(s(t),t)* the arbitrary function) is the function of robot's inner emotional experience.

Let us state that the subject *S(t)* initiates robot's inner emotional experience.

<u>Definition 1.3</u>. The robot's inner emotional experience function *M(t)* is called an 'emotion' if it satisfies the following conditions:
1. Function domain of *M(t)*: $t \in [0, t^0]$, $t^0 > 0$;
2. $t^0 \leq t^*$ (note that this condition is equivalent to emotion termination in case the subject effect is either over or not over yet);
3. *M(t)* is the single-valued function;
4. $M(0) = 0$;
5. $M(t^0) = 0$;
6. *M(t)* is the constant-sign function;
7. There is the derivative $\dfrac{d|M(t)|}{dt}$ within the function domain;

8. There is the only point $z$ within the function domain, such that $z \neq 0$, $z \neq t^0$ and $\left.\dfrac{d|M(t)|}{dt}\right|_{t=z} = 0$;

9. $\dfrac{d|M(t)|}{dt} > 0$ with $t < z$;

10. $\dfrac{d|M(t)|}{dt} < 0$ with $t < z$.

Let us assume there is such $J>0$ that for any emotions of a robot the condition $|M(t)| \leq J$ is valid.

Now we can easily see that the function $M(t) = P\sin\left(\dfrac{\pi}{t^0}t\right)$ for $t \in [0, t^0]$, $P = const$, is an emotion.

<u>Definition 1.4</u>. The function $M(t)$ is called an ambivalent emotion if it can be presented as the vector which elements are emotions initiated simultaneously by one and the same subject.

We will not focus on the content of emotions, and, according to [4], below we plan to take into account only the following things important to us:
1. Emotions have a sign (plus or minus).
2. An object has a *finite number* of emotions.

Based on (2) we conclude that the robot's emotional state can be described by the emotion vector $\overline{M}(t)$ with the finite number of elements (cardinality) equal to $n$:

$$\overline{M}(t) = [M_1(t),...,M_n(t)].$$

Hereinafter, in case we speak about a single-type emotion, we will omit the corresponding index mark, vector mark and will denote this by $M(t)$.

Assume the emotion-free state of a robot as a zero emotion level.

It is obvious, that stimuli can be totally external, partially external (or 'partially memorized') and internal. All of them may become a subject:
- totally external stimuli (which are not contained in the robot's memory (see Fig. 1.1)), may serve as a subject;
-'partially memorized' stimuli (when some part of information about them is entered into the robot's memory, and some part of it comes from the outside as external experience (Fig. 2.2)) may also serve as a subject;
- internal stimuli (when full information about these stimuli is kept in the robot's memory (Fig. 1.3)) may serve as a subject, as well. This is the case when, e.g., some recollection (past event memories) of a robot may generate emotions.

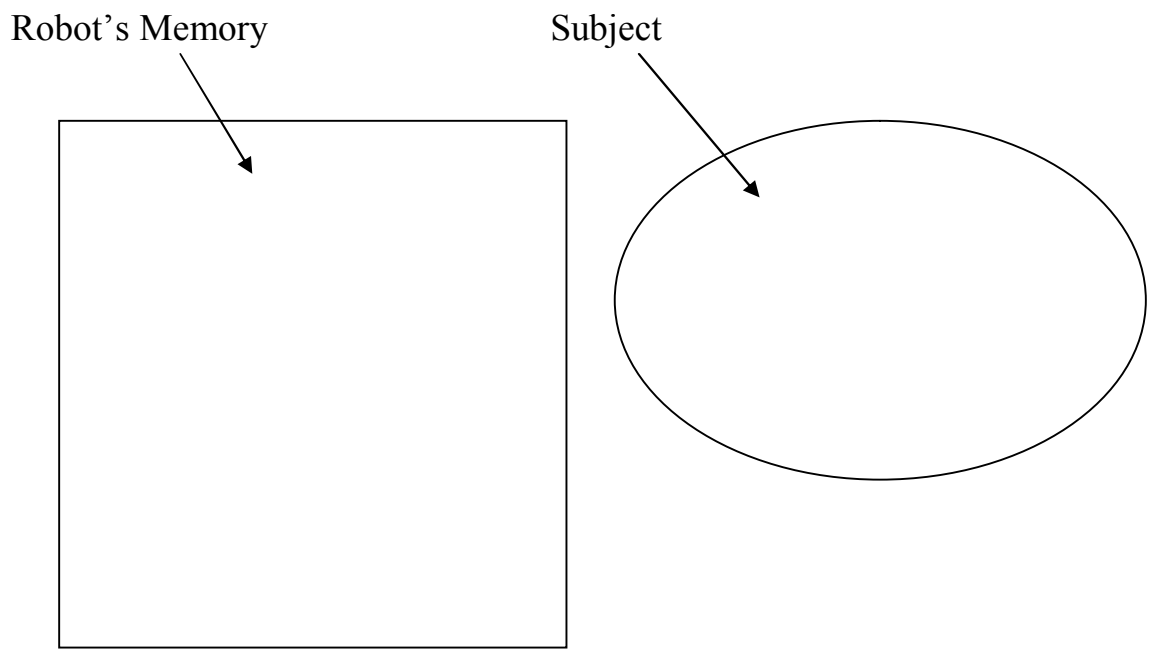

Fig. 1.1. Totally external stimuli as a subject

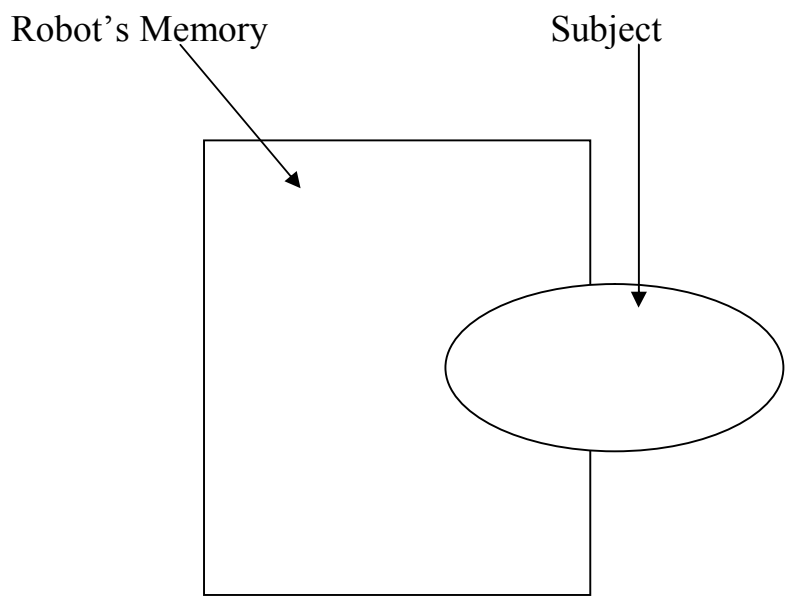

Fig. 1.2. Partially external stimuli as a subject

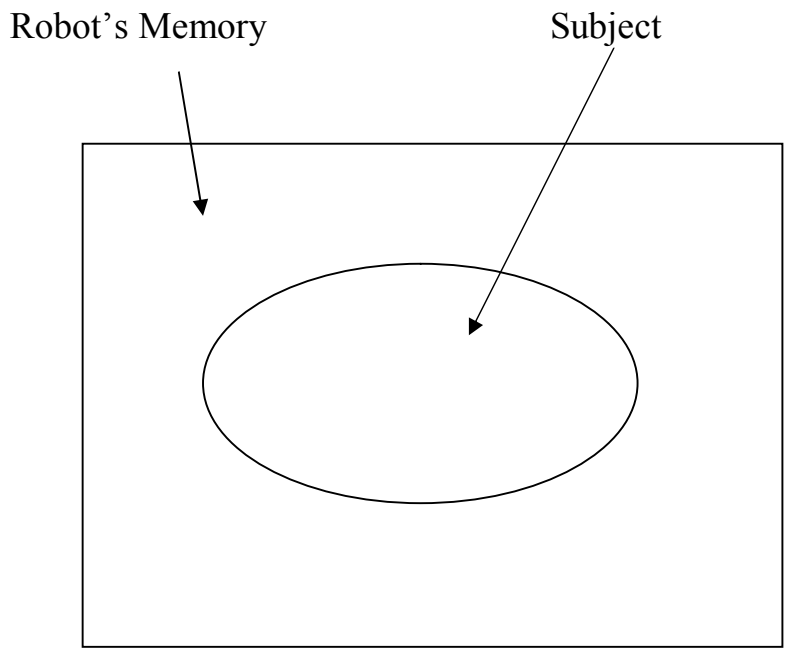

Fig. 1.3. Internal stimuli as a subject

Fig. 1.2 and Fig. 1.3 partially correspond with the psychological theory of S. Schechter [4]. According to Schechter, the occurred emotional state of a person is effected by his/her previous experience and his/her assessment of the current situation, as well as by perceived stimuli and stimulus-initiated physical alterations.

Let us note, that when describing a subject and its belonging to the robot's memory we used the term 'information' which is measured in bits [5]. So, let us advance the following hypothesis: a *subject* can be measured in *bits of information* as well.

It is obvious, that different subjects can initiate one and the same emotion of a robot, i.e. there is no one-to-one dependence between a subject and an emotion (Fig. 1.4).

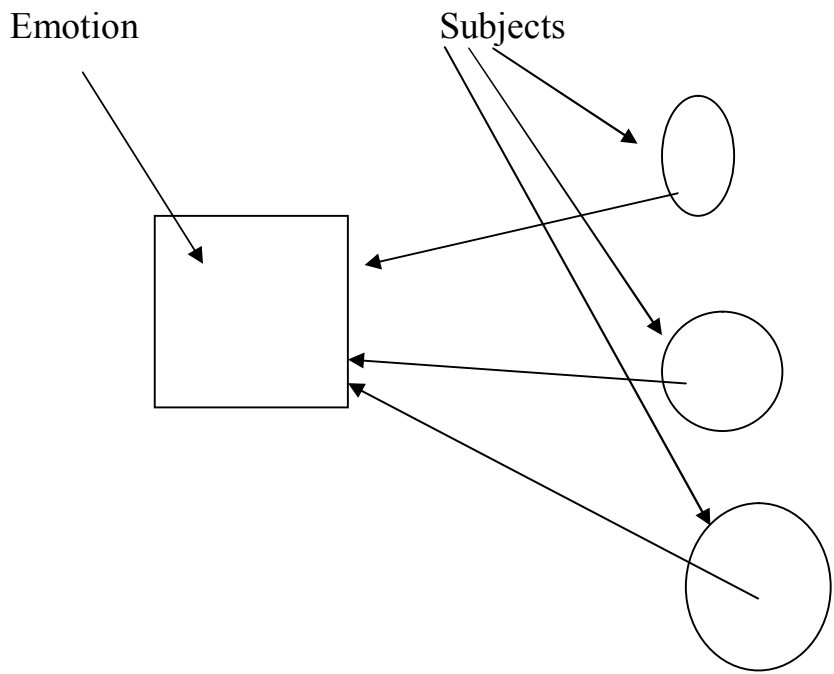

Fig. 1.4. Relation between Subjects and Emotion.

And also, one and the same subject can initiate different emotions of a robot [4] (Fig.1.5).

Let us introduce the concept of the unit (or specific) emotion, similarly to matter density in Physics [6],

Definition 1.5. The specific emotion *a(S(t),t)* of a robot is an emotion per single subject unit.

Obviously, the specific emotion satisfies the following relation:
$$a(S(t),t) = \frac{M(S(t),t)}{S(t)}.$$

We can easily see that the sign of the robot's emotion $M(S(t),t)$ is determined by the sign of the specific emotion *a(S(t),t)*.

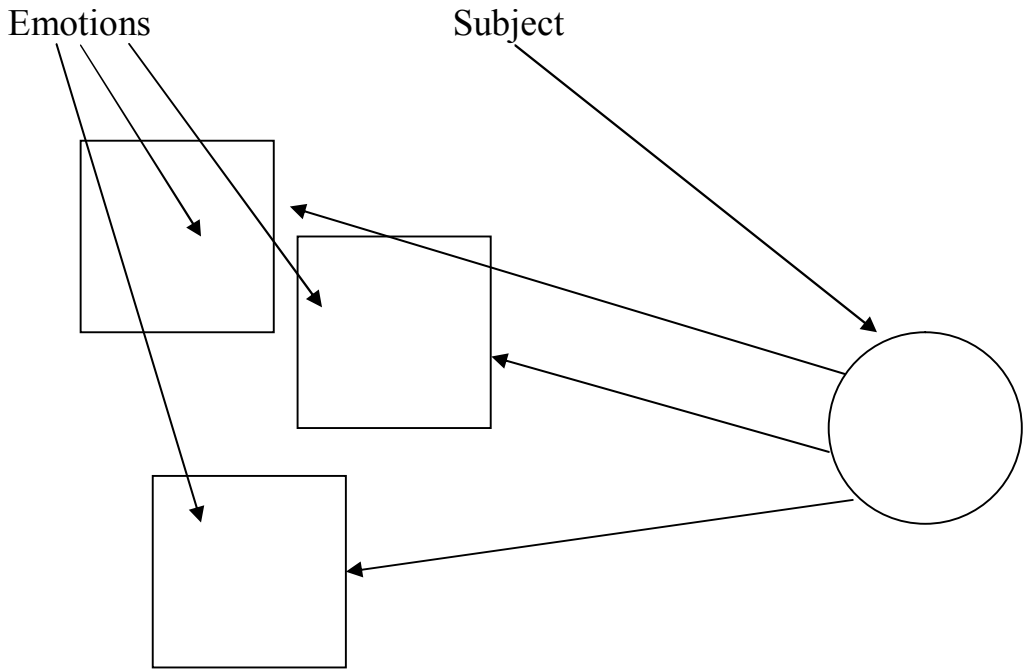

Fig.1.5. Relation between robot's emotions and a subject.

Mathematical theory of emotional robots described in this book considers the cases shown in Fig. 1.4 and 1.5.

## 2. EDUCATION OF A ROBOT

Let us introduce the definition of emotional upbringing (*emotional education*) of a robot abstracting from the psychological concept of education/ upbringing.

<u>Definition 2.1</u>. The *upbringing* or *education* of a robot is a relatively stable attitude of this robot towards a subject.

From Definition 1.3 it follows that the robot's emotion $M(t)$ is the continuous function on the segment $[0, t]$, consequently $M(t)$ is integrable on this segment. Considering that, we can work out the following definition.

<u>Definition 2.2</u>. The elementary robot's education $r(t)$ based on subjects $S(t)$ is the following function:

$$r(t) = \int_0^t a(S(\tau), \tau) S(\tau) d\tau. \qquad (2.1)$$

The obvious mathematical features of the elementary education are as follows:
1) if a specific emotion sign coincides with a subject sign, then the education is positive;
2) in virtue of Definition 2.3, the function $r(t)$ is differentiable with respect to the parameter *t*, so the relation $M(s(t), t) = \dfrac{dr(t)}{dt}$ is valid.

Let us consider that in the course of time a robot can forget emotions experienced some time ago. Its current education is less and less effected by those past (bygone) emotions. Consequently, past elementary educations initiated by those emotions become forgotten as well.

Hence, the following definition becomes obvious.

<u>Definition 2.3</u> The education of a robot *R(t)* based on the subjects *S(t)* is the following function:

$$R(t) = r(\tau) + \theta_i(t) R_i(t_i), \qquad (2.2)$$

where $t$ is the current time, $t > t_i$, $0 \leq \theta_i(t) \leq 1$. The current time satisfies the relation $t = \tau + t_i$, where $\tau$ is the current time of the current emotion effect from the beginning of its initiation, $t_i$ is the total time of all the formerly experienced emotions effect, $R_i(t_i)$ is the education obtained by a robot in the time $t_i$.

A verbal definition of education is as follows: it is a value determining motivation stability of the robot's behavior on a certain class of subjects.

It is obvious, that an education can be measured in bits of information similarly to a subject, and, consequently, emotions are to be measured in bit per second (*bit/s*).

<u>Definition 2.4</u>. Coefficients $\theta_i(t)$ are the memory coefficients of events experienced in the past, i.e. coefficients of the robot's memory.

According to (2.2) we can write down a relation specifying the education in the beginning of the *i*+1[st] emotion effect upon the robot:

$$R_{i+1}(0) = r(0) + \theta_i(0) R_i(t_i).$$

It is easy to see that the eqs.

$$R_{i+1}(0) = R_i(t_i), \quad r(0) = 0.$$

hold true.

Consequently $\theta_i(0) = 1$ is valid.

<u>Definition 2.5.</u> A time step is the effect time of one emotion.

According to results obtained by psychological researches an emotion cannot last more than 10 seconds. Therefore, let us assume that a time step value of any robot emotion is less or equal to 10 sec.

Here and below psychological characteristics of robots corresponding to a current moment of the time step are bracketed after the variable, and psychological characteristics corresponding to the end of time steps are denoted without brackets. For instance, $R_i(t)$ defines a function of education altering for the current time *t* of the valid time step *i*, and $R_i$ defines a value of education in the end of the time step *i*.

It is easy to see that the robot featuring the past event memory coefficient identical with 1 remembers in detail all its past emotional educations. This robot can be regarded as autistic. But let us suppose that the robot's memories of the past events are deleted, i.e. the two-sided inequality $0 \leq \theta_i < 1$ is valid for a forgetful robot in the end of each time step. We are now in position to state a theorem for this kind of robot.

Theorem 2.1. Educating the forgetful robot by means of positive emotions only leads to satiety.
Proof:
It easy to see that Relation (2.2) is equivalent to

$$R(t) = r(\tau) + \theta_i(t)[r_{i-1} + \theta_{i-1}(t)R_{i-2}]. \qquad (2.3)$$

Equation (2.3) can take the form

$$R(t) = r(\tau) + \theta_i r_{i-1} + \theta_i \theta_{i-1} r_{i-2} + \theta_i \theta_{i-1} \theta_{i-2} r_{i-3} + \ldots + \theta_i \theta_{i-1} \theta_{i-2} \ldots \theta_1 r_0. \qquad (2.4)$$

Since all the emotions are positive, elementary educations are positive, too; since all the emotions are value-limited, and time of emotion effect is also limited, so elementary educations are also limited. This makes us conclude that there are $\theta$ and $q$ of a forgetful robot for which the following inequalities hold true:

$$1 > \theta \geq \theta_j, \quad q \geq r_k, \quad q \geq r(\tau), \qquad (2.5)$$

where $j = \overline{1,i}, \quad k = \overline{0,i-1}$.

Due to (2.4) и (2.5) we can obtain the upper bound of the function $R(t)$ variation. It will have the form

$$R(t) \leq q + q \sum_{j=0}^{i-1} \theta^j \leq 2q \sum_{j=0}^{i-1} \theta^j \qquad (2.6)$$

The right side of (2.6) defines the sum of geometric progression terms, which yields inequality

$$R(t) \leq 2q \frac{1 - \theta^{i-1}}{1 - \theta}. \qquad (2.7)$$

Having passed to the limit under $t \to \infty$ or $i \to \infty$ in the right side of (2.7) we get the upper bound of the education value:

$$R(t) \leq \frac{2q}{1 - \theta}. \qquad (2.8)$$

Inequality (2.8) makes us conclude that the robot's education based on positive emotions has the upper bound, i.e. it is satiated.
The proof is now complete.

Psychological researches entirely confirm Theorem 2.1. According to their results, it is not possible to bring up and train a person ad infinitum, as at some certain moment he\she gets satiated [4], and passes to the next stage of his\her emotional activity.

<u>Definition 2.6</u>. The limiting education $U$ is the value corresponding to the end point of emotion effect time and satisfying the relation $U = \dfrac{q}{1-\theta}$.

<u>Definition 2.7.</u> Emotions initiating equal elementary educations are tantamount (equivalent).

<u>Definition 2.8.</u> A uniformly forgetful robot is a forgetful robot whose memory coefficients corresponding to the end point of emotion effect time are constant and equal to each other.

<u>Theorem 2.2</u>. The education $R_i$ based on tantamount emotions of the uniformly forgetful robot is defined by the relation $R_i = q\dfrac{1-\theta^i}{1-\theta}$, where $q$ is the elementary education value, and $i$ is the order number of the initiating tantamount emotion from a quantity of emotions on which basis this education has been being performed by the current time point.
Its <u>proof</u> is evident from Theorem 2.1.

Also let us note the following. When performing a robot's emotion by means of software, it is impossible to predict the subject effect time. Therefore it is expedient to model the emotions after subject effect is over.

Example:
Let us take the emotion function in a form
$$M(t) = P \sin\left[\dfrac{\pi}{t^0 - t^*}(t - t^*)\right], \qquad (2.9)$$
with $P = const$, $t \in [t^*, t^0]$,
$t^0$ the fixed value, at that $t^0 \in (t^*, 2t^*]$.
In (2.9) we replaced conditions 1, 2, 4, 5, 8 in the definition of emotion by the following:
1. Function domain $M(t)$: $t \in [t^*, t^0]$;

2. $t^0 \leq 2t^*$;

4. $M(t^*) = 0$;

5. $M(t^0) = 0$;

8. Function domain contains an only point $z$, such that $z \neq t^*$, $z \neq t^0$ and $\dfrac{d|M(t)|}{dt}\bigg/_{t=z} = 0$.

Also let us note the following: according to (2.9), replacements of several conditions of belonging of the robot's inner emotional experience function $M(t)$ to emotions do not require the currently considered theory to be revised.

Obviously, the time step $\tau$ for Emotion (2.9) satisfies $\tau = t^0 - t^*$, and the elementary education $r$ is computed by

$$r = \int_{t^*}^{t^0} P \sin\left[\frac{\pi}{t^0 - t^*}(t - t^*)\right] dt = 2P \frac{t^0 - t^*}{\pi} = 2P \frac{\tau}{\pi}. \qquad (2.10)$$

We can easily see that during the education process Eq. (2.10) provides tantamount emotions under $\tau = t^0 - t^* = const$.

Let us consider all the time step**s** to be equal to each other.

Below we give a theorem which mathematically characterizes deletion of the past\bygone education memory data if those educations are not maintained by emotions with the course of time. In this case the index $i$ is defined by the relation $i = \left[\dfrac{t}{\sigma}\right]$, with $t$ the current time, $\sigma$ the effect time of the first and only emotion causing the elementary education $r_0$.

Theorem 2.3. The uniformly forgetful robot forgets its first and only elementary education exponentially.

Proof. According to (2.4), if there is no constant emotional effect during some period of time, then the robot's education by the time $t$ satisfies the relation

$$R_i = \theta_i \theta_{i-1} \theta_{i-2} \ldots \theta_1 r_0. \qquad (2.11)$$

As far as the robot is uniformly forgetful, so $\theta_j = \theta = const$, with $j = \overline{1, i}$ is valid. Consequently, $R_i = \theta^i r_0$ holds true.

The proof is now complete.

The next theorem allows assessing the upper bound of the forgetful robot's current education in case when this robot had obtained only one elementary education in the past.

Theorem 2.4. The current education of the forgetful robot obtained due to an only positive elementary education satisfies the inequality $R(t) \leq \theta^{i-1} r_0$, with $\theta \geq \theta_j$, $j = \overline{1,i}$.

Its proof is evident from (2.11).

Above we noted validity of

$$M(t) = \frac{dr(t)}{dt}. \tag{2.12}$$

Assuming that memory coefficients are differentiable functions and taking into consideration (2.12) we get the formula for the sum (*i.e.* resulting) emotion *V(t)*:

$$V(t) = \frac{dr(t)}{dt} + R_{i-1} \frac{d\theta_i(t)}{dt} + \frac{dR_{i-1}(t)}{dt} \theta_i(t). \tag{2.13}$$

(2.13) allows us to assert that sum emotions of the robot depend on past educations, memory coefficients and their rate of change.

It is quite easy to see that for the robot with the absolute emotional memory ($\theta_j \equiv 1$, $j = \overline{1,i}$) current sum emotions are not dependent on past educations.

Let robot's elementary educations satisfy the following inequality:

$$|r_j| \leq q. \tag{2.14}$$

Under *i* tending to infinity and the inverse numeration of elementary educations, (2.4) takes the form:

$$R = \sum_{i=1}^{\infty} r_{i-1} \prod_{j=1}^{i-1} \theta_j. \tag{2.15}$$

Definition 2.9. The robot's education corresponding to (2.15) is an infinite education.

Let us note that the infinite education convergence determines education prospects.

Theorem 2.5. For the forgetful robot, the infinite education corresponding to ends of time steps converges.

Proof. Let us show that Series (2.15) is absolutely convergent.

As $0 \leq \theta_i < 1$ holds true, so there is such $\theta$ less than unity, that $\theta_i \leq \theta < 1$ (with $\overline{i = 1, \infty}$) is valid.

By virtue of Inequality (2.14), Formula (2.15) and formula for finding a sum of terms of a geometric progression [7] we develop a correlation

$$\sum_{i=1}^{\infty} |r_{i-1}| \prod_{j=1}^{i-1} \theta_j \leq \sum_{i=0}^{\infty} q\theta^i = \frac{q}{1-\theta} < \infty.$$

So, Series (2.15) is absolutely convergent, consequently it converges.
The proof is now complete. Quod erat demonstrandum.

By virtue of the theorem given above, the relation
$z = \lim_{i \to \infty} R_i = \lim_{i \to \infty} r_i + \lim_{i \to \infty} \theta_i \lim_{i \to \infty} R_{i-1}$ is valid for the end of each time step of the continuous education process, and this relation is equivalent to

$$z = \lim_{i \to \infty} r_i + \lim_{i \to \infty} \theta_i z. \qquad (2.16)$$

(2.16) allows to enunciate the following theorem.

Theorem 2.6. The uniformly forgetful robot's elementary education corresponding to ends of time steps in the course of continuous education process tends to be constant.

Proof.

As $\theta_i = \theta = const < 1$, $\overline{i = 1, \infty}$,

holds true for the uniformly forgetful robot, by virtue of (3.16) the elementary education sequence corresponding to ends of education time steps, has a limit.

Thus the theorem is proved.

Corollary 2.1. For the uniformly forgetful robot $\lim_{i \to \infty} r_i = (1-\theta)z$ is valid.

The proof follows from (2.16).

Let us assess the extent of error of the infinite education value provided when $k$ terms of series are used for assessing the sum of Series (2.15).

It is easy to see that the inverse numeration of elementary educations makes the error of

$$b_{k+1} = \left| \sum_{i=k+1}^{\infty} r_i \prod_{j=1}^{i-1} \theta_j \right| \text{ satisfy } b_{k+1} \le \frac{q\theta^k}{1-\theta} \text{ under finite summation of } k \text{ terms of}$$

series.

Obviously, an education cannot be performed continuously: after the series of emotional effects there comes a slack period in this education.

Let us introduce a supplementary definition.

<u>Definition 2.10</u>. A complete education cycle is a quantity of time steps equal to the **sum** of time steps under the effect of education emotions and a number of time steps corresponding with the slack period (absence of elementary education effects upon the robot) till the next emotional education effect.

Let us consider the education process of the uniformly forgetful robot with tantamount emotions.

It is easy to see that according to Theorems 2.2 and 2.3 the education $F_{j_1,k_1}$ for the first complete education cycle of the uniformly forgetful robot based on tantamount emotions with equal periods satisfies the following relation:

$$F_{j_1,k_1} = q\theta^{k_1} \frac{1-\theta^{j_1}}{1-\theta}, \qquad (2.17)$$

where $j_1$ is the quantity of time steps in the presence of education effects upon the robot, $k_1$ is the quantity of time steps in their absence.

Obviously, the education $F_{j_n,k_n}$, obtained by the robot as a result of $n$ complete education cycles is determined by the equality

$$F_{j_n,k_n} = \theta^{k_n} \left( q \frac{1-\theta^{j_n}}{1-\theta} + \theta^{j_n} F_{j_{n-1},k_{n-1}} \right). \qquad (2.18)$$

From the forms of Relations (2.17) – (2.18) it follows that $\Omega_{j_n,k_n}$, set by the equality $\Omega_{j_n,k_n} = \frac{F_{j_n,k_n}}{q}$, does not depend on $q$. Since $q=const$ is valid, then $\Omega_{j_n,k_n}$ is a unitless measure for assessing the education obtained by the robot in $n$ complete education cycles.

<u>Definition 2.11</u>. The function $\Omega_{j_n,k_n}$ is a memory function.

It is evident that the memory function shows to what extent tantamount educational emotions are memorized by the robot in the course of the educational process.

Let $U$ defines the value equal to the maximal (satiated) education. Assuming that emotions are tantamount and memory coefficients are equal to one and the same constant, we pass to the limit in both parts of Equality (2.2) under the quantity of time steps tending to infinity. As a result we get

$$\lim_{i \to \infty} r(t) = U(1-\theta) = q.$$

So, the robot's education $R$, obtained in the first complete education cycle is determined by the formula

$$R = \theta^{k_1}\left(1 - \theta^{j_1}\right) U.$$

It easy to see that the function $G(k_1, j_1)$, satisfying the relation

$$G(k_1, j_1) = \frac{R}{U} = \theta^{k_1}\left(1 - \theta^{j_1}\right), \qquad (2.19)$$

determines deviation of the education from its satiety: the closer is $G(k_1, j_1)$ (with the given values $k_1$ and $j_1$) to 1, the closer the robot's education is to its satiety, and vice versa.

<u>Definition 2.12.</u> The function $G(k_1, j_1)$ is a satiety indicator.

It is easy to see that the satiety indicator for the fixed $k_1$ and $j_1$ has a maximum value when the condition

$$\theta = \left(\frac{k_1}{j_1 + k_1}\right)^{\frac{1}{j_1}} \qquad (2.20)$$

holds true.

Inserting (2.20) to Relation (2.19) we get the formula specifying the maximal value $G_{\max}$ of the satiety indicator in the end of the first complete upbringing cycle.

$$G_{\max} = \left(\frac{k_1}{j_1 + k_1}\right)^{\frac{k_1}{j_1}} \frac{j_1}{j_1 + k_1}.$$

<u>Definition 2.13</u>. The function $B_{j_n, k_n} = \dfrac{F_{j_n, k_n}}{U}$ is a complete satiety indicator.

In the conclusion of this chapter we give several statements concerning the non-uniformly forgetful robot with non-tantamount emotions.

It easy to see that for this kind of robot in the end of $n$ complete education cycles the general education function $V^{[n]}_{l_n, i_n}$, defining the education obtained during those cycles, satisfies the relation

$$V^{[p]}_{l_p,i_p} = \left(\prod_{k=1}^{l_p} \widetilde{\theta}^{[p]}_k\right)\left[r^{[p]}_{i_p+1} + \sum_{k=1}^{i_p+1} r^{[p]}_{k-1}\prod_{j=1}^{k}\theta^{[p]}_j + \left(\prod_{i=1}^{i_p}\theta^{[p]}_i\right)V^{[p-1]}_{l_{p-1},i_{p-1}}\right],$$

$p = \overline{2,n}$

$$V^{[1]}_{i_1,l_1} = \left(\prod_{k=1}^{l_1}\widetilde{\theta}^{[1]}_k\right)\left[r^{[1]}_{i_1+1} + \sum_{k=1}^{i_1+1} r^{[1]}_{k-1}\prod_{j=1}^{k}\theta^{[1]}_j\right],$$

where $[i]$ denotes variables corresponding to the $i$-th education cycle, $i = \overline{1,n}$, $\widetilde{\theta}^{[p]}_k$ corresponds to memory coefficients of the $p$-th cycle for time steps without emotional educations, $k$ is the number of the time step without emotional educations, $l_p$ is the quantity of time steps in the $p$-th cycle without emotional effects, $i_p$ is the quantity of time steps in the $p$-th education cycle with continuous emotional education effects.

Obviously, for the forgetful robot the following inequalities are valid:

$$|V^{[p]}_{l_p,i_p}| \leq F_{l_p,i_p}, \quad F_{l_p,i_p} = \theta^{l_p}\left(q\frac{1}{1-\theta} + \theta^{i_p} F_{l_{p-1},i_{p-1}}\right) \quad p = \overline{2,n},$$

$$\left|V^{[1]}_{l_1,i_1}\right| \leq F_{l_1,i_1}, \quad F_{l_1,i_1} = q\theta^{i_1}\frac{1}{1-\theta},$$

where $\theta = \max(\widetilde{\theta}^{[p]}_j, \theta^{[p]}_i)$, $i = \overline{1,i_p}$, $j = \overline{1,l_p}$, $p = \overline{1,n}$.

Let us introduce the following definition.

<u>Definition 2.14.</u> The generalized memory function $W^{[n]}_{l_n,i_n}$ is a value satisfying the relation $\widetilde{W}^{[n]}_{l_n,i_n} = \dfrac{V^{[n]}_{l_n,i_n}}{q}$.

<u>Definition 2.15.</u> The generalized education satiety indicator is the function

$$W^{[n]}_{l_n,i_n} = \frac{|V^{[n]}_{l_n,i_n}|(1-\theta)}{q}.$$

Based on the definitions given above we conclude that the generalized memory function and the generalized education satiety indicator are unitless functions.

It is obvious that the generalized education satiety indicator satisfies the inequality $0 \leq W^{[n]}_{l_n,i_n} \leq 1$.

# 3. PARAMETERS OF A GROUP OF EMOTIONAL ROBOTS

Let us consider a problem connected with studying emotional conditions of the group of robots. The theory given below represents one of attempts to formalize mathematically the solution of this problem.

<u>Definition 3.1</u>. The sum (*i.e.* resulting) education of the group including $n$ robots belonging to the set $\Omega_n$ based on the subject $S(t)$ is computed as follows:

$$W_{\Omega_n} = \sum_{i \in \Omega_n} \int_0^t a_i(S(\tau), \tau) S(\tau) d\tau. \qquad (3.1)$$

Suppose we have two groups including $p$ and $k$ robots and forming two sets $\Omega_p$, $\Omega_k$ correspondingly, where $\Omega_p \cup \Omega_k = \Omega_n$, $\Omega_p \cap \Omega_k = \otimes$, $\Omega_p \neq \otimes$, $\Omega_k \neq \otimes$. Let us find out when the utmost psychological conflict between those groups can occur on one and the same class of subjects. It is obvious that, for instance, hatred (odium) is determined by opposite-signed sum educations of rival groups; also it is obvious that the equality $\dfrac{W_{\Omega_k}}{W_{\Omega_p}} = -1$ (where $W_{\Omega_p} \neq 0$) is to hold true so that the utmost confrontation between robot groups become possible.

The *converse proposition* is valid:

If a sum education of two groups is equal to zero and an education of at least one robot is nonzero, then the utmost confrontation is most likely possible between two groups of robots.

Below we give the proof of this statement:
Suppose $W_{\Omega_n} = 0$, then $k$ and $p$ can be selected so that $k+p=n$, as well as $\Omega_k$ and $\Omega_p$ can be selected so that $W_{\Omega_n} = W_{\Omega_k} + W_{\Omega_p} = 0$ is valid, i.e. $\dfrac{W_{\Omega_k}}{W_{\Omega_p}} = -1$ under $W_{\Omega_p} \neq 0$, which required to be proved.

Based on this we get <u>Theorem 3.1</u>. The necessary and sufficient condition for the utmost confrontation between robot groups including at least one robot with a nonzero education is that the sum education of those groups equals to zero.

Obviously, the farther is $\left| W_{\Omega_k} \right|$ from zero, the worse is the confrontation.

The given theorem helps us to define the most rival pairs of robots or robot groups. To find out the pairs of rival groups it is enough to calculate each robot education and then obtain a set of all possible sum educations (e.g., by enumerative technique, manually or by computer). Sets of robots with sum educations close to zero make up rival risk groups.

It is easy to see that the greater the sum education of a group differs from zero, the more united (or, better say, more serried) this group is.

Suppose the sum education of members of the first group obtained in the course of several complete education cycles $W^{[1]}$ satisfies the relation $W^{[1]} = \sum_{j=1}^{n} V^{[1]}_{i^{[1]}_{p_j}, k^{[1]}_{k_{p_j}}}$, and the corresponding sum education of the second group is computed by the formula $W^{[2]} = \sum_{j=1}^{m} V^{[2]}_{i^{[2]}_{p_j}, k^{[2]}_{k_{p_j}}}$, where the index [1] or [2] denotes belonging to Group 1 or Group 2, $n$ is a quantity of robots in Group 1, $m$ is a quantity of robots in Group 2.

Then the condition of rivalry between those groups is defined by the relation $W^{[1]} + W^{[2]} = 0$, which is equivalent to

$$\sum_{j=1}^{n} V^{[1]}_{i^{[1]}_{p_j}, k^{[1]}_{k_{p_j}}} + \sum_{j=1}^{m} V^{[2]}_{i^{[2]}_{p_j}, k^{[2]}_{k_{p_j}}} = 0.$$

<u>Definition 3.2</u>. Re-education (re-bringing) is change of the education sign to the opposite one.

Obviously, Group 1 including $k$ robots can re-educate Group 2 including $p$ robots in its favour if the equality $\dfrac{W_{\Omega_k}}{W_{\Omega_p}} = Q$ where $Q \neq -1$, $|W_{\Omega_k}| > |W_{\Omega_p}|$, $W_{\Omega_p} W_{\Omega_k} < 0$ holds true by the beginning of the re-educating process. The greater $Q$ differs from -1, the more effective is this re-education.

<u>Definition 3.3</u>. There is an emotional conflict in the group at the time $t_0$ if the sum of emotions of each member in the group is equal to zero, i.e. $\sum_{i=1}^{n} M_i(t_0) = 0$.

Obviously, if at the time $t_0$ sum emotions and educations of members of the group are equal to zero, then there is the open conflict threat at its utmost stage.

Let us consider conditions of the conflict between uniformly forgetful robots with tantamount emotions.

According to the definitions given above, the limiting education of the first uniformly forgetful robot $U_1$ educated by tantamount emotions, satisfies the relation $U_1 = \dfrac{q_1}{1-\theta_1}$, and the limiting education of the second tantamount emotions, is defined by uniformly forgetful robot $U_2$ also educated by the relation $U_2 = \dfrac{q_2}{1-\theta_2}$ where $\theta_1$ and $\theta_2$ are memory coefficients, $q_1$ and $q_2$ are values of the corresponding elementary educations. Suppose in the course of an infinite education process robots come to an education conflict. This implies that the formula $U_1 = U_2$ is valid, and so is the relation

$$\frac{q_1}{1-\theta_1} = \frac{q_2}{1-\theta_2}. \qquad (3.2)$$

Equality (3.2) allows us to compute the approximate interdependence of memory coefficients of two uniformly forgetful robots conflicting on tantamount emotions:

$$\theta_2 = 1 - (1-\theta_1)\frac{q_2}{q_1}. \qquad (3.3)$$

It is obvious, that if coefficients $\theta_1$ and $\theta_2$ are not connected by Relation (2.6), then Robot 1 and Robot 2 will never come to an education conflict at the limit.

Above in Chapter 2 we showed that in the course of $j$ continuous education effects on Robot 1 and $i$ continuous education effects on Robot 2 the corresponding educations can be described as

$$R_j^{[1]} = q_1 \frac{1-\theta_1^j}{1-\theta_1}, \quad R_i^{[2]} = q_2 \frac{1-\theta_2^i}{1-\theta_2}.$$

Then the condition of the onset of the conflict in the education process can be computed by the equality

$$q_1 \frac{1-\theta_1^j}{1-\theta_1} = q_2 \frac{1-\theta_2^i}{1-\theta_2}. \qquad (3.4)$$

But we can state that if memory coefficients $\theta_1$ and $\theta_2$ are not connected by Relation (3.3), then the conflict between robots ceases with time by itself, i.e. without any extra emotional effects different from already existing emotion effects.

## 4. FRIENDSHIP BETWEEN ROBOTS: FELLOWSHIP (CONCORDANCE)

This chapter represents an attempt to introduce the term and concept of "friendship between robots", which we prefer to characterize as fellowship or concordance of robots.

Here we introduce a couple of definitions.

<u>Definition 4.1</u>. The group of robots is a united *fellowship* if individual educations of each member are positive.

Definition 4.2. If individual educations of a fellowship are not less than $P_0 > 0$, then $P_0$ is the *fellowship value* of this group.

Theorem 4.1. There exists $\xi$ such that a fellowship value of a fellowship is $\xi$.

Proof: As this group of robots is a fellowship, then individual educations $\overline{R_i}$ $(i=1,n)$ of each member satisfy the condition $R_i > 0$. Therefore there exists a value $\xi>0$ such that the inequalities $R_i \geq \xi$, $i = \overline{1,n}$ hold true.
This completes the proof of the Theorem.

Definition 4.3. Suppose individual educations of a group including *n* robots are positive. A sum (total) fellowship value of *n* robots is a sum of all individual education values of robots in this group.

Assume that a set of *n* robots is divided into two sub-groups. Suppose the first Sub-group including *m* robots is more united and affinitive of the two fellowships, and its fellowship value is $P_0$. So, the sum/total fellowship value of the first Sub-group *P* is computed by the equality $P = mP_0$.

Assume the second Sub-group includes *n-m* robots and has a fellowship value $R_0$. Then the sum/total fellowship value of the first Sub-group *A* is defined by the equality $A = (n-m)R_0$.

Obviously, the sum/total fellowship value *R* of two sub-groups is defined by the formula
$$R = P + A = mP_0 + (n-m)R_0. \qquad (4.1)$$

Assume the inequality $P_0 > R_0$ holds true.

Suppose members of the second Sub-group are robots with equal tantamount emotions *q* and uniformly forgetful with equal memory coefficients $\theta$.

We state the following problem: let us define the education condition for robots of the second Sub-group, under which it is possible for the fellowship coefficient of the second Sub-group to become equal or more than the fellowship coefficient of the first Sub-group as a result of education of robots in the second Sub-group.

Based on (4.1) we conclude that this condition is determined by the inequality

$$mP_0 + (n-m)R_* \geq nP_0, \qquad (4.2)$$

where $R_*$ is the education value of each robot in the second sub-group after the education process had started.

It is easy to see that Relation (4.2) is equivalent to the formula
$$R_* \geq P_0. \qquad (4.3)$$

Let us effect simultaneously on each robot of the second sub-group by tantamount emotions until Condition (4.3) becomes to hold true. Obviously, by the end of the education process the relation

$$q\frac{1-\theta^j}{1-\theta}+\theta^j R_0 \geq P_0$$

is to hold true, where $j$ is a quantity of education process time steps for robots of the second sub-group.

So, for finding the least quantity (number) of the necessary education time steps with the given memory coefficients of robots of the second sub-group we are to solve the following problem:

solve for

$$\min_{j\geq 1}\left(q\frac{1-\theta^j}{1-\theta}+\theta^j R_0 - P_0\right) \qquad (4.4)$$

under

$$q\frac{1-\theta^j}{1-\theta}+\theta^j R_0 - P_0 \geq 0.$$

Let us prove the theorem.

Theorem 4.2. If the relation $\dfrac{q}{1-\theta}+R_0 < P_0$ is valid, then Problem (4.4) has no solution.

Proof. Since robots in the second sub-group are uniformly forgetful, then the two-sided inequality $0 \leq \theta < 1$ holds true. So, Theorem 4.2 statement yields a formula valid for any time step value $j$:

$$q\frac{1-\theta^j}{1-\theta}+\theta^j R_0 < P_0$$

This formula implies that the limiting condition in Problem (4.4) is never to hold true. Therefore, this task has no solution under this theorem statement.
This completes the proof.

In other words, the theorem implies the following: "education effects not necessarily make robots achieve equal fellowship (i.e. concordance) between members of the group with the given fellowship value".

## 5. EQUIVALENT EDUCATION PROCESSES

Definition 5.1. The equivalent education process is a continuous education process corresponding to an education with tantamount emotions, equal memory coefficients and featuring the minimal deviation at all the education assessment node points from the values of a real continuous education process of a robot.

## 5.1. MATHEMATICAL MODEL OF EQUIVALENT EDUCATION PROCESSES

Suppose education values of a real continuous process are established in the end of each period by values $R_j$, $j = \overline{1, n}$, where $n$ is a total quantity of education time steps. Also suppose conditions

$$R_{j+1} \geq R_j > 0,\ j = \overline{1, n-1}. \tag{5.1}$$

are valid.

Now we approximate the real education process to an equivalent education process. To do this we need to find such $\theta, q$ under which the objective function reaches its minimum

$$J(\theta, q) = \sum_{j=2}^{n} \left( R_j - \theta^{j-1} R_1 - q \frac{1 - \theta^{j-1}}{1 - \theta} \right)^2. \tag{5.2}$$

So, in order to develop the equivalent education process we need to solve the equation set

$$\frac{\partial J(\theta, q)}{\partial \theta} = 0,\quad \frac{\partial J(\theta, q)}{\partial q} = 0. \tag{5.3}$$

Considering Relation (5.2), Equation set (5.3) in its expanded version takes the form:

$$\sum_{j=2}^{n} \left( R_j - \theta^{j-1} R_1 - q \frac{1 - \theta^{j-1}}{1 - \theta} \right)\left(1 - \theta^{j-1}\right) = 0, \tag{5.4}$$

$$\sum_{j=2}^{n} \left( R_j - \theta^{j-1} R_1 - q \frac{1 - \theta^{j-1}}{1 - \theta} \right)\left[ (j-1)\theta^{j-2} R_1 - q \frac{1 - \theta^{j-1} - (j-1)\theta^{j-2}(1-\theta)}{(1-\theta)^2} \right]$$

$$\tag{5.5}$$

Since for adequately selected time steps the solutions of Equation sets (5.4) – (5.5) have to satisfy the conditions

$$0 \leq \theta < 1,\ q \geq 0, \tag{5.6}$$

then, due to checking on validity of (5.6) we can estimate adequacy of the equivalent process to the real education process.

Coefficients $\theta, q$ solved out of Eqs. (5.4) – (5.5) allow us to find approximately the limiting value of the education of the continuous process $Z$. Obviously, $Z$ satisfies the relation

$$Z = \lim_{j \to \infty} \left( R_1 \theta^{j-1} + q \frac{1 - \theta^{j-1}}{1 - \theta} \right) = \frac{q}{1 - \theta}.$$

Let us assess the error in calculations of the limiting education in the real continuous process through the equivalent education process.

According to the formula of continuous education in the real process, the relation

$$R_j = r_j + \theta_j R_{j-1} + \prod_{k=1}^{j} \theta_k R_1. \tag{5.7}$$

holds true.

In (5.7) we pass to the limit under the time tending to infinity:

$$\lim_{j \to \infty} R_j = \lim_{j \to \infty} r_j + \lim_{j \to \infty} \theta_j \lim_{j \to \infty} R_{j-1} + \lim_{j \to \infty} \prod_{k=1}^{j} \theta_k R_1. \tag{5.8}$$

According to the theorem of education convergence, the relation $\lim_{j \to \infty} R_j = D > 0$ holds true. Hence, Relation (5.8) is tantamount to the equality

$$D = \lim_{j \to \infty} r_j + \lim_{t \to \infty} \theta_j D.$$

So the value $D$ satisfies the relation

$$D = \frac{\lim_{j \to \infty} r_j}{1 - \lim_{j \to \infty} \theta_j}. \tag{5.9}$$

Suppose the inequality

$$\frac{\lim_{j \to \infty} r_j}{1 - \lim_{j \to \infty} \theta_j} \geq \frac{q}{1 - \theta} \tag{5.10}$$

holds true.

Considering the last inequality and Relation (5.9) we get the following formula:

$$D - Z = \frac{\lim_{j \to \infty} r_j}{1 - \lim_{j \to \infty} \theta_j} - \frac{q}{1 - \theta} \leq \frac{M}{1 - \bar{\theta}} - \frac{q}{1 - \theta} = \frac{M - q}{(1 - \bar{\theta})(1 - \theta)} + \frac{q\bar{\theta} - M\theta}{(1 - \bar{\theta})(1 - \theta)}, \tag{5.11}$$

where $M = \max_j r_j$, $\bar{\theta} = \max_j \theta_j$.

Let us consider the case corresponding to the inequality $\dfrac{\lim\limits_{j\to\infty} r_j}{1-\lim\limits_{j\to\infty}\theta_j} < \dfrac{q}{1-\theta}$.

Obviously, in this case the limiting education error estimate satisfies the relations

$$Z - D = \frac{q}{1-\theta} - \frac{\lim\limits_{j\to\infty} r_j}{1-\lim\limits_{j\to\infty}\theta_j} \leq \frac{q}{1-\theta} - \frac{M}{1-\underline{\theta}} = \frac{q(1-\underline{\theta}) - M(1-\theta)}{(1-\theta)(1-\underline{\theta})}, \qquad (5.12)$$

where $M = \min r_j$, $\underline{\theta} = \min \theta_j$, $j = \overline{1,\infty}$.

Relations (5.11) and (5.12) allow us to get the error estimate $X$ of the limiting education under approximation of the real process to the equivalent education process. Obviously, in the general case it can be found by the formula

$$X \leq \max\left(\left|\frac{M-q}{(1-\underline{\theta})(1-\theta)} + \frac{q\underline{\theta} - M\theta}{(1-\underline{\theta})(1-\theta)}\right|, \left|\frac{q(1-\underline{\theta}) - M(1-\theta)}{(1-\theta)(1-\underline{\theta})}\right|\right).$$

Analyzing Formulas (5.11) and (5.12) we can state that the worse is the robot's emotional memory the less is the error estimate of the limiting education.

Also, (5.11) and (5.12) allow us to state that the formula

$$\lim_{j\to\infty} R_j \approx \frac{q}{1-\theta} \qquad (5.13)$$

holds true if the matter concerns a forgetful robot.

By virtue of (5.1), Relation (5.13) allows us to find approximately the limiting education of a robot for the real educating process on the basis of the equivalent educating process.

It is easy to see that (5.9) implies the relation

$$R_j \leq \frac{M}{1-\underline{\theta}}, \quad j = \overline{1,\infty}$$

which is the upper bound of the education value of the forgetful robot's real education process.

## 5.2. ALTERNATIVE TO AN OBJECTIVE FUNCTION UNDER COINCIDENCE OF TIME STEPS OF REAL AND EQUIVALENT EDUCATION PROCESSES

Let us introduce a simpler objective function such that its minimization can give us the coefficients $\theta$ and $q$ which define the equivalent education process

$$J(\theta, q) = \sum_{i=2}^{n} (R_i - q - \theta R_{i-1})^2.$$

Validity of this objective function for designing an equivalent education process follows from the formula of education of a robot with tantamount emotions and equal memory coefficients: $R_i = q + \theta R_{i-1}$.

In order to minimize this function let us solve the following equation set:

$$\begin{cases} \dfrac{\partial J(\theta, q)}{\partial \theta} = 0, \\ \dfrac{\partial J(\theta, q)}{\partial q} = 0. \end{cases}$$

Now we are to find the corresponding derivatives:

$$\begin{cases} \dfrac{\partial J(\theta, q)}{\partial \theta} = 2\sum_{i=2}^{n} (R_i - q - \theta R_{i-1})(-R_{i-1}), \\ \dfrac{\partial J(\theta, q)}{\partial q} = 2\sum_{i=2}^{n} (R_i - q - \theta R_{i-1})(-1). \end{cases}$$

Then the system takes the form

$$\begin{cases} \sum_{i=2}^{n} (R_i - q - \theta R_{i-1}) R_{i-1} = 0, \\ \sum_{i=2}^{n} (R_i - q - \theta R_{i-1}) = 0. \end{cases}$$

Now simplify this and get

$$\begin{cases} \sum_{i=2}^{n} R_i R_{i-1} - q \sum_{i=2}^{n} R_{i-1} - \theta \sum_{i=2}^{n} (R_{i-1})^2 = 0, \\ \sum_{i=2}^{n} R_i - q(n-1) - \theta \sum_{i=2}^{n} R_{i-1} = 0. \end{cases}$$

The system is linear relative to $\theta$ and $q$ so let's express $\theta$ and $q$ as $R_i$. Out of the second equation we get

$$q = \dfrac{\sum_{i=2}^{n} R_i - \theta \sum_{i=2}^{n} R_{i-1}}{n-1}.$$

Substitution of $q$ into the first equation gives

$$\sum_{i=2}^{n} R_i R_{i-1} - \frac{\sum_{i=2}^{n} R_i - \theta \sum_{i=2}^{n} R_{i-1}}{n-1} \sum_{i=2}^{n} R_{i-1} - \theta \sum_{i=21}^{n}(R_{i-1})^2 = 0 \Leftrightarrow$$

$$\Leftrightarrow \sum_{i=2}^{n} R_i R_{i-1} - \frac{\sum_{i=2}^{n} R_i \sum_{i=2}^{n} R_{i-1}}{n-1} + \theta \frac{\left(\sum_{i=2}^{n} R_{i-1}\right)^2}{n-1} - \theta \sum_{i=2}^{n}(R_{i-1})^2 = 0 \Leftrightarrow$$

$$\Leftrightarrow \sum_{i=2}^{n} R_i R_{i-1} - \frac{\sum_{i=2}^{n} R_i \sum_{i=2}^{n} R_{i-1}}{n-1} - \theta \left(\sum_{i=2}^{n}(R_{i-1})^2 - \frac{\left(\sum_{i=2}^{n} R_{i-1}\right)^2}{n-1}\right) = 0,$$

$$\theta = \frac{\sum_{i=2}^{n} R_i R_{i-1} - \frac{\sum_{i=2}^{n} R_i \sum_{i=2}^{n} R_{i-1}}{n-1}}{\left(\sum_{i=2}^{n}(R_{i-1})^2 - \frac{\left(\sum_{i=2}^{n} R_{i-1}\right)^2}{n-1}\right)} = \frac{(n-1)\sum_{i=2}^{n} R_i R_{i-1} - \sum_{i=2}^{n} R_i \sum_{i=2}^{n} R_{i-1}}{(n-1)\sum_{i=2}^{n}(R_{i-1})^2 - \left(\sum_{i=2}^{n} R_{i-1}\right)^2}.$$

Consequently,

$$q = \frac{\sum_{i=2}^{n} R_i - \frac{(n-1)\sum_{i=2}^{n} R_i R_{i-1} - \sum_{i=2}^{n} R_i \sum_{i=2}^{n} R_{i-1}}{(n-1)\sum_{i=2}^{n}(R_{i-1})^2 - \left(\sum_{i=2}^{n} R_{i-1}\right)^2} \sum_{i=2}^{n} R_{i-1}}{n-1}$$

So, under known education values of the real education process of a robot $R_i, i = \overline{1,n}$ we get unique values of $\theta$ and $q$ for which the conditions $0 \leq \theta < 1, q \geq 0$ are to be valid.

If the obtained values satisfy all the limitations mentioned above, then the coefficients $\theta$ and $q$ define the equivalent education process. If the obtained values do not satisfy those limitations, then it is not possible to develop any equivalent education process with the same time steps as in the real education process and with the corresponding educations $R_i, i = \overline{1,n}$ of the real education process.

The obtained coefficients $\theta$ and $q$ allow us to find approximately the limiting value of the real education process. Let $Z$ be the limiting value; then

$$Z = \lim_{i \to \infty}(q + \theta R_{i-1}) = q + \theta Z.$$

Out of this we get $Z = \dfrac{q}{1-\theta}$.

According to the formula of the continuous education process the relation
$$R_i = r_i + \theta_i R_{i-1}.$$

is valid.

Having passed to the limit in this relation we get
$$\lim_{i \to \infty} R_i = \lim_{i \to \infty} r_i + \lim_{i \to \infty} \theta_i \lim_{i \to \infty} R_{i-1}.$$

According to Theorem 2.1 of forgetful robot's education convergence at positive emotions, the relation $\lim_{i \to \infty} R_i = D > 0$ holds true. Hence, we get

$$D = \lim_{i \to \infty} r_i + \lim_{i \to \infty} \theta_i D,$$

$$D = \frac{\lim_{i \to \infty} r_i}{1 - \lim_{i \to \infty} \theta_i}.$$

Let $\dfrac{\lim_{i \to \infty} r_i}{1 - \lim_{i \to \infty} \theta_i} \geq \dfrac{q}{1 - \theta}$; then we get the following formula:

$$D - Z = \frac{\lim_{i \to \infty} r_i}{1 - \lim_{i \to \infty} \theta_i} - \frac{q}{1 - \theta} \leq \frac{M_1}{1 - \bar{\theta}_1} - \frac{q}{1 - \theta} = \frac{M_1(1 - \theta) - q(1 - \bar{\theta}_1)}{(1 - \bar{\theta}_1)(1 - \theta)},$$

with: $M_1 = \max_i r_i, \bar{\theta}_1 = \max_i \theta_i, i = \overline{1, \infty}$

Let us consider the case when $\dfrac{\lim_{i \to \infty} r_i}{1 - \lim_{i \to \infty} \theta_i} < \dfrac{q}{1 - \theta}$, out of it we get the following formula:

$$Z - D = \frac{q}{1 - \theta} - \frac{\lim_{i \to \infty} r_i}{1 - \lim_{i \to \infty} \theta_i} \leq \frac{q}{1 - \theta} - \frac{M_2}{1 - \bar{\theta}_2} = \frac{q(1 - \bar{\theta}_2) - M_2(1 - \theta)}{(1 - \bar{\theta}_2)(1 - \theta)},$$

where $M_2 = \min_i r_i, \bar{\theta}_2 = \min_i \theta_i, i = \overline{1, \infty}$

The obtained relations are necessary for computing an error of the limiting education under approximation of the real education process to the equivalent education process. The error $X$ is found by

$$X \leq \max\left( \frac{M_1(1 - \theta) - q(1 - \bar{\theta}_1)}{(1 - \bar{\theta}_1)(1 - \theta)}, \frac{q(1 - \bar{\theta}_2) - M_2(1 - \theta)}{(1 - \bar{\theta}_2)(1 - \theta)} \right).$$

Analyzing the inequality described above we conclude that the worse is the robot's emotional memory, the less is the error of limiting education computing.

<u>Example</u>. Let us consider an example of equivalent education process development.

Suppose the real education process includes three education time steps $R_1, R_2, R_3$ with $R_1 = 1, R_2 = 3, R_3 = 4$. By the formulas given above we find $\theta$ and $q$, and get

$$\theta = \frac{2*15 - 7*4}{2*10 - 16} = \frac{1}{2} = 0.5$$

$$q = \frac{7 - 0.5*4}{2} = \frac{5}{2} = 2.5$$

At that, $0 \leq \theta < 1, q \geq 0$ are valid.

So, we obtained an approximation of the real education process including three time steps with the real education $R_1 = 1, R_2 = 3, R_3 = 4$ to the equivalent education process with tantamount emotions under $q = 2.5$ and equal memory coefficients $\theta = 0.5$.

Based on the obtained values, we can find the approximate value of the limiting education $Z$. Simple calculations lead to the following relation: $Z \approx \frac{q}{1-\theta} = 5$.

### 5.3. GENERALIZATION IN CASE OF NONCOINCIDENCE OF TIME STEPS OF REAL AND EQUIVALENT EDUCATION PROCESSES

Speaking about generalization, assume that a number of education time steps in the equivalent education process may differ from their number in the real education process. For instance, the end of the second time step of the real education process may coincide with the end of the second or more time step of the equivalent education process.

Noncoincidence of time steps for education processes can occur due to randomness in timing of educations of the real education process. Education values of the real process can be approximately restored for each time step in the course of development of the equivalent education process.

Assuming that the equivalent education process is continuous, we can suppose that during each time step our robot is effected by a tantamount emotion with the elementary education $q$.

It is easy to see that the objective function can be presented as follows:

$$J(\theta, q, j_1, ..., j_n) = \sum_{i=1}^{n} \left( R_i - q \frac{1 - \theta^{j_i}}{1 - \theta} \right)^2, \quad (5.14)$$

where $R_i$ is the education value of the real education process after the time step $i$, and $q \frac{1 - \theta^{j_i}}{1 - \theta}$ characterizes the education obtained as a result of the equivalent education process after the time step $j_i$.

So, in order to develop the equivalent education process it is necessary to minimize Objective function (5.14). For that we need to solve the following equation set:

$$\frac{\partial J(\theta, q, j_1, \ldots, j_n)}{\partial \theta} = 0,$$

$$\frac{\partial J(\theta, q, j_1, \ldots, j_n)}{\partial q} = 0.$$

Then the equation set for finding will take a form

$$\begin{cases} \sum_{i=1}^{n} \left( R_i - q \frac{1 - \theta^{j_i}}{1 - \theta} \right) \left[ j_i \theta^{j_i - 1}(1 - \theta) + \theta^{j_i} - 1 \right] = 0, \\ \sum_{i=1}^{n} \left( R_i - q \frac{1 - \theta^{j_i}}{1 - \theta} \right) = 0, \\ 0 < \theta < 1, q > 0 \end{cases}$$

Example.

Assuming that $R_1 = 3, R_2 = 6, R_3 = 10$ hold true and applying the cyclic data search method for Objective function (5.14) minimization with an enumeration step equal to 0.1 for $q$ and $\theta$, and an enumeration step equal to 1 for $j_i$, and with variation intervals of $q$ between 0.1 and 2.9, $\theta$ - between 0.09 до 0.99. $j_i$ - between 1 до 100, we get the following values: $q = 0.2$, $\theta = 0.99$. $j_1 = 16$, $j_2 = 35$, $j_3 = 69$. Obviously, the limiting education equals 20. The computation results show that under found parameters of the equivalent education process the value of (5.14) equals 0.0056, i.e. the developed equivalent education process approximates the real one quite closely.

## 6. METHOD OF APPROXIMATE DEFINITION OF MEMORY COEFFICIENT FUNCTION

In Chapter 2 we proved the following equality for the beginning of each time step:

$$\theta_i(0) = 1, i = \overline{1, \infty}. \tag{6.1}$$

Now let us express the memory coefficients $\theta_i(t)$ in the following form

$$\theta_i(t) = a_i t + b_i,$$

where $a_i$, $b_i$ are constants which are not dependent on the current time $t$ of emotion effect.

According to (6.1) and relations for finding the coefficients $a_i$, $b_i$ we can work out the following equations system:

$$a_i 0 + b_i = 1, \tag{6.2}$$

$$a_i(t_i - t_{i-1}) + b_i = \theta \tag{6.3}$$

with $t_{i-1}, t_i$, the time of the beginning of the $i$-th time step; $\theta$, the memory coefficient of the equivalent process.

We obtain relations allowing us to find the unknown values in Equation system (6.2) – (6.3) provided that parameters of the equivalent process are found on the basis of

The objective function given in Section 5.2.

It is easy to see that the sought-for values are found by the explicit formulas $b_i = 1$,

$$a_i = \frac{\dfrac{(n-1)\sum\limits_{i=2}^{n} R_i R_{i-1} - \sum\limits_{i=2}^{n} R_i \sum\limits_{i=2}^{n} R_{i-1}}{(n-1)\sum\limits_{i=2}^{n} (R_{i-1})^2 - \left(\sum\limits_{i=2}^{n} R_{i-1}\right)^2} - 1}{t_i - t_{i-1}},$$

where $n$ is the number of time steps for which successive values of the robot's education $R_i$ are known, as well as time step which are defined by the values $t_{i-1}, t_i$, $i = \overline{1, n}$.

## 7. MATHEMATICAL MODEL OF FORMING TANTAMOUNT ROBOT SUB-GROUPS

This chapter describes one of the ways to make up groups of robots with equal sum educations.

Let us consider a group of $k$ robots, where each robot has its order number $i$, where $i = \overline{1, k}$.

Suppose the robot $i$ has its education $R_i$. Then the sum education of the group of robots $A$ satisfies the relation $A = \sum\limits_{i=1}^{k} R_i$.

<u>Problem</u>: Out of the set $\Omega$ including all the robots, let us make up sub-groups which are nonoverlapping subsets $\Omega_p$, $p = \overline{1, n}$ ($n < k$), $\bigcup\limits_{p=1}^{n} \Omega_p = \Omega$, so that sum education values of the obtained sub-groups are least different from each other.

Let us give the following definition and prove the auxiliary theorem.

<u>Definition 7.1.</u> The average education $F_p$ of the group $p$ is a value satisfying the relation $F_p = \dfrac{\sum\limits_{j \in \Omega_p} R_j}{N_p}$, where $N_p$ is the quantity of robot units in the set $\Omega_p$.

<u>Theorem 7.1.</u> The sum education $A$ satisfies the equality $A = \sum\limits_{i=1}^{n} N_i F_i$.

<u>Proof.</u> It is easy to see the validity of the equality chain

$$N_i F_i = N_i \dfrac{\sum\limits_{j \in \Omega_i} R_j}{N_i} = \sum\limits_{j \in \Omega_i} R_j . \qquad (7.1)$$

Summing (7.1) with respect to all the values $i$ we get

$$\sum\limits_{i=1}^{n} N_i F_i = \sum\limits_{i=1}^{n} \sum\limits_{j \in \Omega_i} R_j = \sum\limits_{s=1}^{k} R_s = A, \text{ i.e. } \sum\limits_{i=1}^{n} N_i F_i = A.$$

The proof is complete.

Let us introduce the objective function in a form:

$$J = \sum\limits_{i=1}^{n-1} \sum\limits_{j=i+1}^{n} \left( N_i F_i - N_j F_j \right)^2 .$$

Now the problem put above can be mathematically described as follows: solve for

$$\min_{N_i, F_i} J\left(\overline{N}, \overline{F}\right) \qquad (7.2)$$

under limits

$$\sum\limits_{i=1}^{n} N_i = k, \quad \sum\limits_{i=1}^{n} N_i F_i = A, \quad N_i > 0, \quad i = \overline{1, n}.$$

Problem (7.2) deals with determination of conditional extremum of function of several variables, so it can be easily solved by the well-known Lagrange method.

As a result of applying the Lagrange method to the solution of this Problem we get the roots of the following equation system:

$$2F_i \sum\limits_{j=i+1}^{n} \left( N_i F_i - N_j F_j \right) - \lambda_1 - \lambda_2 F_i = 0, \quad i = \overline{1, n-1}, \quad \sum\limits_{i=1}^{n} N_i - k = 0,$$

$$2 \sum\limits_{i=1}^{n-1} \left( N_i F_i - N_n F_n \right) + \lambda_2 = 0, \qquad (7.3)$$

$$\sum_{i=1}^{n} N_i F_i - A = 0, \quad 2\sum_{j=i+1}^{n}(N_i F_i - N_j F_j) - \lambda_2 = 0, \quad i = \overline{1, n-1},$$

$$2F_n \sum_{i=1}^{n-1}(N_i F_i - N_n F_n) + \lambda_1 + \lambda_2 F_n = 0,$$

where $\lambda_1$, $\lambda_2$ are the Lagrange method auxiliary variables.

In the general case, the question about existing and uniqueness of the solution of the nonlinear algebraic equation set (7.3), and about mathematical ways of its solution is still open-ended.

Now let us consider the task which is a little bit different, though similar to Problem (7.2) in its statement. In this new problem statement we suppose that the quantity of robots $N_p$ in the groups $\Omega_p$ is already predetermined. It is quite easy to see that in this case the mathematical statement of the problem will have the following form:

solve for

$$\min_{F_i} J(\overline{F}) \tag{7.4}$$

under $\sum_{i=1}^{n} N_i F_i = A$

According to the Lagrange method, Problem (7.4) solution is reduced to just finding the roots of the linear equation set

$$2\sum_{j=i+1}^{n}(N_i F_i - N_j F_j) - \lambda = 0, \quad i = \overline{1, n-1},$$

$$\sum_{i=1}^{n} N_i F_i - A = 0, \tag{7.5}$$

$$2\sum_{i=1}^{n-1}(N_i F_i - N_n F_n) + \lambda = 0,$$

where $\lambda$ is the Lagrange method auxiliary variable.

It is easy to show that the major equation determinant in this equation set is nonzero (e.g., the case when $n = 2$ means that the group is split into two sub-groups), i.e. with such $n$ this set always has a unique solution.

<u>Definition 7.2</u>. Sub-groups with the values $F_i$, $i = \overline{1, n}$ obtained in the solution of Problem (7.4) are tantamount ones.

<u>Definition 7.3</u>. Sub-groups with the values $F_i$, $i = \overline{1, n}$ which are the solution of Problem (7.4) and which make the objective function $J$ reach its minimum equal to zero are absolutely tantamount sub-groups.

Let us define simple conditions under which the sub-groups being formed are absolutely tantamount.

The minimum of the function $J\left(\bar{F}\right)$ obviously equals to zero when the relations

$$N_i F_i = N_j F_j, \quad i = \overline{1, n-1}, \quad j = \overline{i+1, n},$$

$$\sum_{i=1}^{n} N_i F_i = A.$$

hold true.

It is easy to see that under $n = 2$ the sub-groups become absolutely tantamount when the relations

$$F_1 = \frac{A}{2N_1}, \quad F_2 = \frac{A}{2N_2} \quad \text{hold true.}$$

The solution of Problem (5.9) allows us to get numerical values of abstract average educations which may not coincide with real average educations of sub-groups being formed. This is connected with the fact that average educations of all real sub-groups are known values and, consequently, absolutely tantamount sub-groups might not be obtained basing on educations of single robot units. This is also the reason why it is not always possible to split a set of robots into tantamount sub-groups.

## 8. ALGORITHM FOR FORMING TANTAMOUNT SUB-GROUPS OF ROBOT

Below we give an algorithm for making up real robot sub-groups closest to tantamount ones:

1. Set up values $N_1, ..., N_n$ determining a quantity (number) of robots in each sub-group being formed, with $\sum_{i=1}^{n} N_i = k$.

2. Make up the array $Z$ of different sets $Z = \left\{\Omega_{N_1, y}, ..., \Omega_{N_n, y}\right\}_{y=1}^{q}$ ($q$ is the quantity of set pools in the array $Z$), such that $\bigcup_{i=1}^{n}\Omega_{N_i, y} = \Omega, \ \Omega_{N_i, y} \cap \Omega_{N_j, y} = \varnothing, \ i \neq j, \ i = \overline{1, n}, \ j = \overline{1, n}.$

3. Based on Step 2 find the value of the function $J\left(\bar{F}\right)$ for each pool of sets $\Omega_{N_1, y}, ..., \Omega_{N_n, y}.$

4. Define numbers of $y$ for which the corresponding sets make the objective function $J(\bar{F})$ reach its minimum.

5. Arrange a visual output of sets $\Omega_{N_1,y},...,\Omega_{N_n,y}$, corresponding to the minimum values of $J(\bar{F})$.

Note that performing Step 2 on a computer one may use well-known computer algorithms of combinatory analysis given in [8].

Having selected the sets including robot sub-groups with the closest sum educations, we can assess their equivalence, i.e. to what extent those sub-groups are tantamount towards each other, by comparing average educations of those sub-groups to the values $F_i$ which are the solution result of Problem (7.4).

For assessing the closeness $V$ of the formed sub-groups to the tantamount ones, we suggest applying the following formula:
$V = \max_i \frac{|D_i - F_i|}{F_i}$, $i = \overline{1,n}$, $D_i$ are real average educations of each of formed sub-group. Obviously, the nearer is $V$ to zero, the closer are the formed sub-groups to each other.

To detect sub-groups of robots grouped according to their education levels out of a general set we suggest applying well-known algorithms of cluster analysis [9]. These algorithms may, for instance, help to detect either robots belong to leading or lagging sub-groups.

## 9. APPLYING VECTOR ALGEBRA RULES TO INVESTIGATION OF ROBOT SUB-GROUP EMOTIONAL STATE

Here and below we use Cartesian rectangular coordinates.

<u>Definition 9.1</u>. A robot's education based on $n$ emotion types is the vector $\bar{R} = (R_1, R_2, ..., R_j, ..., R_n)$, where each element of the vector of education based on single-type emotions is defined according to Relation (2.2).

Introducing vectors of educations and emotions allows us to use rules of vector algebra in mathematical operations with educations and emotions.

Thus, the group education $R$ including $m$ robots can be found by the formula

$$R = \sum_{k=1}^{m} \bar{R}_k, \quad (9.1)$$

and the group emotion $M$ can be found by

$$M = \sum_{k=1}^{m} \overline{M}_k, \qquad (9.2)$$

where *k* is an order number of a robot in its group.

Note that with *m<n* the vector of group emotions includes at least *n-m* zero elements.

By introducing Relations (9.1) and (9.2) we obtained a rule for composition of robots' psychological characteristic vectors.

Below we give the results of theoretical research concerning a pair of emotional robots or their two groups; either of the groups features its education and emotion vector.

Definition 9.2. A single-type psychological vector of a robot is either just an emotion vector or just an education vector.

To unify the records we designate single-type psychological vectors as $\overline{a}$ è $\overline{b}$.

Let us consider psychological properties of scalar product of emotion education vectors.

Suppose $\overline{a}$ is the single-type psychological vector of the first robot, or the group of robots, and $\overline{b}$ is the single-type psychological vector of the second robot, or the second group of robots (both of robots or the groups belong to a common set).

According to the rules of vector algebra, a scalar product of two single-type psychological vectors is a value satisfying the relation

$$\left(\overline{a},\overline{b}\right) = \left|\overline{a}\right|\left|\overline{b}\right|\cos\alpha,$$

with:

$\left|\overline{a}\right|, \left|\overline{b}\right|$ the moduli of vectors, obtained by well-known vector algebra formulas;

*α* the vectorial angle contained by $\overline{a}, \overline{b}$.

It is obvious that $\cos(\alpha)$ satisfies the equality

$$\cos(\alpha) = \frac{\left(\overline{a},\overline{b}\right)}{\left|\overline{a}\right|\left|\overline{b}\right|}$$

<u>Definition 9.3</u>. If $\alpha \in \left[0, \dfrac{\pi}{2}\right)$ then we consider that psychological effects are directed at achieving one goal; but if $\alpha \in \left(\dfrac{\pi}{2}, \pi\right]$, then these effects are directed at achieving opposite goals.

Definition 9.3. is illustrated by Fig. 9.1 and Fig. 9.2.

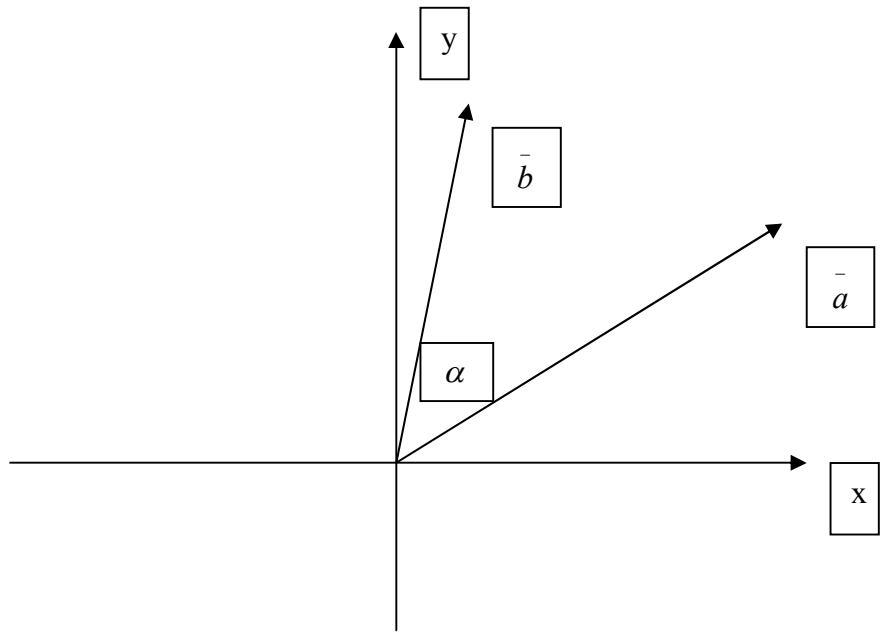

Fig. 9.1. Single-type psychological vectors are directed at achieving one goal

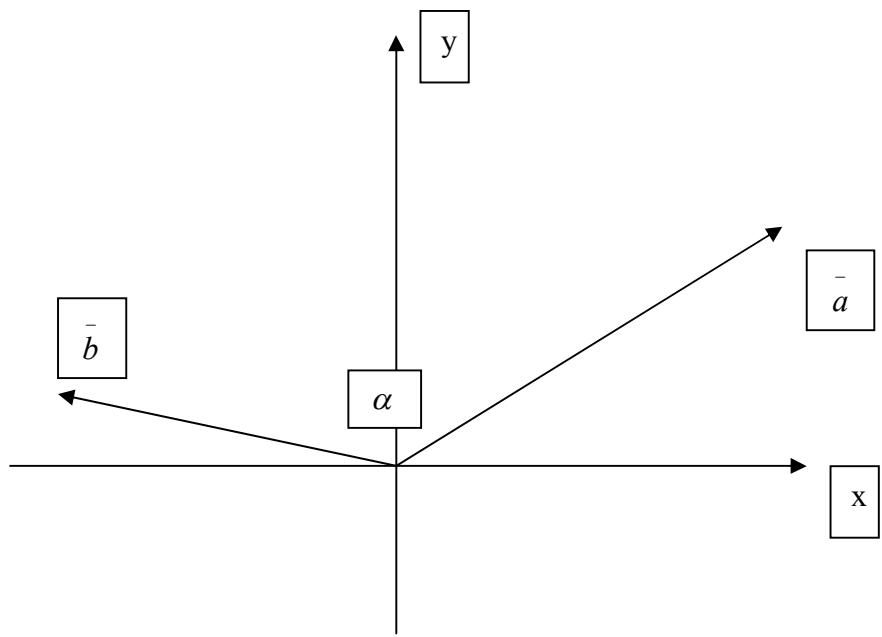

Fig. 9.2. Single-type psychological vectors are directed at achieving different (opposite) goals

The following statements are obvious.

Theorem 9.1. If a cosine of the angle between two single-type psychological vectors is positive, then psychological effects are directed at achieving one common goal.

Corollary 9.1. If a cosine of the angle between two single-type psychological vectors is equal to 1, then psychological effects directed at achieving one goal are the most effective.

Theorem 9.2. If a cosine of the angle between two single-type psychological vectors is negative, then psychological effects contradict each other and are directed at achieving different (opposite) goals.

Corollary 9.2.1. If a cosine of the angle between two single-type psychological vectors is equal to -1, then the set of robots contains sub-groups with opposite psychological characteristics.

Corollary 9.2.2. If a cosine of the angle between two single-type psychological vectors is equal to -1, and moduli of these vectors are equal to each other, then there is a conflict in the set of robots, and psychological characteristics of this conflict correspond to the considered psychological vectors.

It is obvious, that if Corollary 9.2.2. is valid simultaneously for both emotion psychological vectors and education psychological vectors, then the conflict between two sub-groups gets its acutest form. Thus we can formulate the following theorem.

Theorem 9.3. If cosines of the angles between emotion vectors and education vectors are equal to -1, moduli of emotion vectors are equal to each other, and moduli of education vectors are equal to each other, too, then there is a conflict in its peak point.

Theorem 9.4. If a cosine of the angle between single-type psychological vectors is equal to zero, then there occurs an unstable psychological situation so that any single emotion may tend the set of robots either to one goal achieving or to different (opposite) goals achieving (i.e. either to serrying together and uniting or to dissociating and disuniting).
The proof is obvious.

Theorem 9.5. A set of emotional robots can not simultaneously be in situations when the cosine modulus of angle between single-type psychological vectors is equal to 1, and, at the same time, the set's psychological situation is unstable due to this type of vectors.

Proof. Assume the set of robots is emotional. Then its single-type psychological vector is not equal to zero.

Since the cosine modulus of the angle between psychological vectors is equal to 1, then the vectors are collinear. As the collection of robots experiences an unstable psychological situation, so psychological vectors are orthogonal. But both described cases can simultaneously be valid only if at least one vector is equal to zero, but it contradicts with an assumption that the considered set of robots is emotional. So, the theorem *is proved* ex contrario.

Corollary 9.5. Theorem 9.5. can be rephrased as follows: a collection of emotional robots cannot simultaneously experience a conflict and psychological uncertainty.

## 10. MATHEMATICAL ASSESSMENT OF GOAL ACHIEVMENT EXTENT

Suppose an educator set for a robot a numerically expressed goal of education. In some cases it is possible to assess numerically to what extent the robot manages to reach its goal in the course of this education.

### 10.1. Rule of solving for the extent of goal achievement

Let us introduce the following definitions.

Definition 10.1. A goal is the vector $A = (a_1,...,a_m)$ characterizing the desired final state of a robot, achieved in K steps, with $\sum_{i=1}^{m} a_i^2 > 0$.

Below we consider the case when for achieving the goal we have a given number of steps K.

Definition 10.2. A step-to-the goal k is the vector $R_k = (r_{k,1},...,r_{k,m})$ defining a state of a robot obtained in one k-th step in the course of achieving the goal.

Definition 10.3. A state vector (*or* robot's state vector) $W_k$ is a vector corresponding to the goal achievement as a result of passing all the steps through the step k inclusive, and satisfying the relation $W_k = \sum_{i=1}^{k} R_i$.

Obviously, deviation of the step k direction from the goal direction is characterized by the angle $\beta_k$ equal to the angle between the goal itself and the step k to the goal. The cosine of this angle is defined by the formula [10]

$$\cos(\beta_k) = \frac{(A, R_k)}{|A||R_k|}, \tag{10.1}$$

and the cosine of the angle $\alpha_k$ contained by the robot state vector and the goal (this cosine characterizes deviation from the goal direction after passing through $k$ steps) is defined by the relation

$$\cos(\alpha_k) = \frac{(A, W_k)}{|A||W_k|}. \tag{10.2}$$

After passing through the given number of steps $K$ specified for this goal achievement, it is possible to find the value $\delta$ showing how close the robot is to the preset goal. A formula defining $\delta$ is a ratio of the vector projection numerical value $W_K$ onto the vector $A$ to the modulus of $A$.

So, considering (10.2), the relation for evaluating $\delta$ takes a form

$$\delta = \frac{|W_K|\cos(\alpha_K)}{|A|} = \frac{|W_K|}{|A|} \frac{(A, W_K)}{|A||W_K|} = \frac{(A, W_K)}{|A|^2}. \tag{10.3}$$

It is easy to see that $\delta$ can possess any values, and the goal is achieved completely if $\delta \geq 1$.

The cosine of the angle of deviation of the sum state vector from the goal direction $\psi$ can be found by the relation

$$\cos(\psi) = \frac{(A, W_K)}{|A||W_K|}. \tag{10.4}$$

The formula for evaluating the goal achievement percentage $\chi_k$ at every step $k$ is analogous:

$$\chi_k = \frac{(A, R_k)}{|A|^2}, \tag{10.5}$$

and the goal achievement $\lambda_k$ after passing through $k$ steps is found by

$$\lambda_k = \frac{(A, W_k)}{|A|^2}. \tag{10.6}$$

Suppose $t_k$ is time necessary for performing the step $k$, then we can evaluate the total period of time $T$ spent for achieving $\delta$. It is found by the formula

$$T = \sum_{k=1}^{K} t_k.$$

Comparing members of one sub-group to each other we can find the most talented for education robot according to the following criterion: while his positive $\delta$ is equal to that value of the rest robots, he must perform the least time $T$.

The formulas mentioned above can be used for analyzing robot's actions while achieving the goal: for instance, if under some $k$ the values $\chi_k$ are too big (see (10.5)) and angles $\beta_k$ are close to zero (see (10.1)), we may conclude that the robot's actions chosen at the step $k$ provide the most successful achievement of the preset goal.

Obviously, successful actions of the robot at every $k$-step yield the biggest $\delta$, $\lambda_k$ (see (10.3), (10.6)) and the angle values $\psi$, $\alpha_k$ (see (10.4), (10.2)) close to zero. In other words, to achieve the goal successfully the robot must perform maximal results at every step.

Now let us consider a question about quantitative assessment of a sub-group goal achievement.

Suppose every member $j$ of a sub-group has its individual goal $z_j = (h_{j,1},...,h_{j,m})$, where $j = \overline{1,L}$, $L$ is a quantity of robots in the sub-group.

In this case the general goal $A$ of the sub-group is evaluated by
$$A = \sum_{j=1}^{L} z_j.$$

Suppose every robot in the sub-group has its $k$-step to its own goal defined by the vector $f_{j,k} = (S_{j,1,k},...,S_{j,m,k})$, then the sum $k$-step of the sub-group achieving the goal is evaluated by the formula $R_k = \sum_{j=1}^{L} f_{j,k}$, and in $k$ steps the state vector of the sub-group will satisfy the relation $W_k = \sum_{i=1}^{k} R_i = \sum_{i=1}^{k} \sum_{j=1}^{L} f_{j,i}$.

Now, based on the relations introduced, we may numerically assess the achievement extent of the sub-group goal by formulas initially developed for a single robot unit by substituting a robot for a sub-group.

Suppose, having achieved some goal a robot sets up another one. That new goal may have a quantity of components different from the previous one. To find the quantitative assessment of the next goal achievement we can apply the scheme described above including there a corresponding quantity of components of a new goal.

But sometimes the goal of robot's actions cannot be seen clearly. In this case the best way to present this goal is by the matrix $A$:
$$A = \begin{pmatrix} a_{1,1} & ... & a_{1,m} \\ ... & ... & ... \\ a_{q,1} & ... & a_{q,m} \end{pmatrix},$$
where every line represents one of the goals.

Assessing one by one every line (*i.e.* every goal achievement) in the matrix *A* after passing *K* steps we can find the goal which is achieved best of all the rest. Solution of this problem can warn the robot against aiming at unaccomplishable and unrealizable goals.

Also let us pay attention at a simpler case when the goal and *k*-steps are scalars. Note, that in this case the goal and *k*-step to that goal have only two directions – either coinciding with the number axis direction or opposing it. Thus Relations (10.3), (10.5), (10.6) take the form:

$$\delta = \frac{W_K}{A}, \quad \chi_k = \frac{R_k}{A}, \quad \lambda_k = \frac{W_k}{A},$$

where *A* is a value of the goal.

The method of individual assessment of a goal achievement can be used for ranking robots according to their educations, e.g. in descending order. For correct ranking, it is necessary, first of all, to set up the maximally possible accomplishable goal, and then get robots ranked according to numerical values of this goal achievement. If these numerical values appear to be equal for some robots, then a robot with the least deviation from the goal direction has to be put at the first place. This way of education ranking we call goalizing.

Let us consider the case when numerical values of goal vector elements are unknown, but the task is to rank education vectors according to the achievement ascending order (i.e. according to the order of closeness to the goal being achieved). Without breaking the integrity, suppose the robot's goal is to obtain the best result. Then the goal *A* can be characterized by a vector with *m* unit elements: $A = (1,...,1)$. Having assigned an order number to each unit element of the education vector set (this is to indicate its closeness to 1), we get the vector $B_j = (b_{1,j},...,b_{m,j}), \quad j = \overline{1,n}$ for each education.

It is easy to see that in this case the values of projections $\delta_j$ of each vector $B_j$ onto the goal vector *A* satisfy the relation

$$\delta_j = \frac{\sum\limits_{i=1}^{m} B_{i,j}}{\sqrt{m}}, \qquad (10.7)$$

and the angle of deviation from the goal achievement $\Psi_j$ can be found by the formula

$$\cos \Psi_j = \frac{\sum\limits_{i=1}^{m} B_{i,j}}{\sqrt{\sum\limits_{i=1}^{m} B_{i,j}^2} \sqrt{m}}.$$

According to (10.7), the less is $\delta_j$, the closer are the vectors $B_j$ to the goal. Thus these vectors can be ranked in ascending order of $\delta_j$. If with all this $\delta_i = \delta_k$, $i \neq k$, then the vector corresponding to the biggest $cos \Psi_j$ is to be put forward.

Let us note the following.
Sometimes a robot achieves its final goal stepwise, from one part of the goal to the other.
Suppose the final goal is evaluated by the vector
$$\overline{A} = (a_1,...,a_{k_1}, a_{k_1+1},...,a_{k_j},...,a_{k_{j+1}}, a_{k_{j+1}+1}...,a_{k_{j+1}},...,a_{k_n+1},...,a_m),$$
where $n$ is a number of elements of the finite education goal vector.
Without breaking the integrity, suppose that at the step $i$ the robot achieved the education
$$\overline{W}_i = (R_1,...,R_{k_1},...,R_{k_i},...,R_{k_{i+1}}, 0,...,0).$$

Then in (10.3) the vector $W_i$ satisfies the relation $W_i = \overline{W}_i$, where $i = \overline{1,s}$, $s$ is the total amount of steps to the goal.

## 10.2. Algorithm for forming tantamount sub-groups of robots according to their goal achievement extent

Based on the rule of solving for the extent of goal achievement given above in Section 10.1 we can suggest the following algorithm of forming two tantamount sub-groups, if goals of each robot are equal to each other and each sub-group includes an even number of members:

1) Make up a general linear array out of goal achievement extent values of each robot;

2) Within the array, define numbers of robots having the maximal and minimal values of goal achievement extent;

3) Robots with these numbers go to the first sub-group;
4) Remove from the general array the elements with maximal and minimal values of goal achievement extent;

5) If the resulting general array is not empty, then go to Step 6, otherwise go to Step 10;
6) Within the resulting general array, define numbers of robots having the maximal and minimal values of goal achievement extent;
7) Robots with these numbers go to the second sub-group;

8) Remove from the general array the elements with maximal and minimal values of goal achievement extent;

9) If the resulting general array is not empty, then go to Step 2, otherwise go to Step 10;

10) End.

## 11. MATHEMATICAL MODEL OF ROBOT's EMOTIONAL ABILITIES

In the previous section we presented formulas for evaluating the extent of education goal achievement based on methods of the vector ranking projective theory.

Now we advance a hypothesis that the ablest 'gifted' robot is the most docile and submissive to education, i.e. by the time $t$ this robot reaches large average extent of education goal achievement per time unit. On the basis of this hypothesis we offer a relation for evaluating the robot's ability $F$:

$$F(t) = \frac{d\left(\frac{\int_0^t \delta(\tau)d\tau}{t}\right)}{dt} = \frac{\frac{\sum_{i=1}^m a_i R_i(t)}{\sum_{i=1}^m a_i^2} t - \frac{\int_0^t \sum_{i=1}^m a_i R_i(\tau)d\tau}{\sum_{i=1}^m a_i^2}}{t^2}. \qquad (11.1)$$

So, robot's abilities are measured in units reciprocal of the time.

Based on Chapter 3 we can get the assessment of abilities of a robot which does not have a property of absolute memory. This assessment is given by

$$|F(t)| \leq \frac{4q \sum_{i=1}^m |a_i| \frac{1-\theta_i^{j_i}}{1-\theta_i}}{t \sum_{i=1}^m a_i^2},$$

with:

$q = \max_i |r_i|$;

$\theta_i$ the values of maximal memory coefficients corresponding to the $i$-th education;

$j_i$ the order number of the $i$-th education time step depending on the education current time $t$.

Let us prove the following theorem.

Theorem 11.1. The abilities $F_k$ of the forgetful robot are limited at the end of each time step $k$.

Proof. Suppose $\theta = \max\limits_{i=1,n} \theta_i$, $\tau$ is the minimum value of all the periods. Then the inequality

$$|F_k| \leq \frac{4q \sum_{i=1}^{m} |a_i| \frac{1-\theta_i^{j_i}}{1-\theta_i}}{t_k \sum_{i=1}^{m} a_i^2} \leq \frac{4 \frac{q}{1-\theta} \sum_{i=1}^{m} |a_i|}{\tau \sum_{i=1}^{m} a_i^2}$$

holds true, quod erat demonstrandum.

Eq. (11.1) finds the most capable ('gifted') robot in a group, ranks robots according to their abilities and discloses robots with highly pronounced propensities to this or that scope of activities defined by subsets of education vector elements.

We offer the following algorithm for finding scopes of activities to which a robot has the strongest abilities.

1. Set up the general education goal vector $A = (a_1,...,a_m)$ as an input parameter.

2. By the control point of time $t$ the education process is to result in the general education vector $R = (R_1(t),...,R_m(t))$.

3. Select subgoal vectors (which are subsets composed of one, two, …, $m$ elements of the goal vector) in series from the goal vector $A$.

4. Compute the ability values for each of these composed subsets provided that the considered educations correspond to numbers of elements of subgoal vectors.

5. Select the maximal ability values corresponding to each of composed subsets.

6. Find numbers of elements of the composed subset goals, corresponding to these maximal ability values. These numbers correspond to education types according to which a robot is considered to be the most successful, i.e. the ablest.

The relation $N = \sum_{i=1}^{n} C_n^i$ defines the quantity of major steps $N$ to be performed when this algorithms is processed by computer software.

For studying robot's abilities it is necessary to introduce the concept of ability range implying the quantity of educations matching the given ability value. We can conclude that, with equal ability values, the wider is the robot's ability range, the more talented is the robot. Thus, general abilities of a robot are defined by $B$, satisfying the equality $B = (p, F)$, where $p$ is the ability range, $F$ is the ability value.

Theorem 11.2. In a univariate case, with the time infinitely increasing the abilities of a forgetful robot achieving the goal tend to zero.
Proof. Since the relations

$$V^{[p]}_{l_p,i_p} = \left(\prod_{k=1}^{l_p} \tilde{\theta}^{[p]}_k\right)\left[r^{[p]}_{i_p+1} + \sum_{k=1}^{i_p+1} r^{[p]}_{k-1} \prod_{j=1}^{k} \theta^{[p]}_j + \left(\prod_{i=1}^{i_p} \theta^{[p]}_i V^{[p-1]}_{l_{p-1},i_{p-1}}\right)\right], \quad p = \overline{2,n},$$

$$V^{[1]}_{i_1,l_1} = \left(\prod_{k=1}^{l_1} \tilde{\theta}^{[1]}_k\right)\left[r^{[1]}_{i_1+1} + \sum_{k=1}^{i_1+1} r^{[1]}_{k-1} \prod_{j=1}^{k} \theta^{[1]}_j\right] \quad (11.2)$$

are valid for $n$ education cycles, then the inequalities

$$|V^{[p]}_{l_p,i_p}| \le F_{l_p,i_p},$$

$$F_{l_p,i_p} = \theta^{l_p}\left(q\frac{1}{1-\theta} + \theta^{i_p} F_{l_{p-1},i_{p-1}}\right),$$

$p = \overline{2,n}$,

$$\left|V^{[1]}_{l_1,i_1}\right| \le F_{l_1,i_1}, \quad F_{l_1,i_1} = q\theta^{i_1}\frac{1}{1-\theta}, \quad (11.3)$$

with $\theta = \max(\tilde{\theta}^{[p]}_j, \theta^{[p]}_i)$, $i = \overline{1,i_p}$, $j = \overline{1,l_p}$, $p = \overline{1,n}$

hold true for the forgetful robot.

Also, Formulas (11.3) imply the chain of relations
$$\left|V^{[n]}_{l_n,i_n}\right| \le \sum_{i=1}^{n+1}\theta^{i-1}q\frac{1}{1-\theta} \le \sum_{i=1}^{\infty}\left(q\frac{1}{1-\theta}\right)\theta^{i-1} = \frac{q}{(1-\theta)^2}. \quad (11.4)$$

In view of the definition of the robot's ability for the univariate case the following formulas can be written:

$$\lim_{t\to\infty}|Z(t)| = \lim_{t\to\infty}\left|\frac{d\left(\frac{\int_0^t V_{l_n,i_n}^{[n]}(\tau)d\tau}{|A|t}\right)}{dt}\right| \le \lim_{t\to\infty}\frac{4\frac{q}{(1-\theta)^2}t}{|A|t^2} = \lim_{t\to\infty}\frac{4\frac{q}{(1-\theta)^2}}{|A|t} = 0.$$

So, $\lim_{t\to\infty} Z(t) = 0$. Thus the theorem is proved.

Theorem 11.3. In a multivariate case, with the time tending to infinity the abilities of a forgetful robot tend to zero.

Proof. Since for each education component $j$ the values $\left|V_{l_{n_j},i_{n_j}}^{[n_j]}\right|$ satisfy the relations $|V_{l_{n_j},i_{n_j}}^{[n_j]}| \le F_{l_{n_j},i_{n_j}}$, (where $j=\overline{1,m}$, $m$ is the number of goal vector components and current education vector components, $n_j$ is the number of complete education cycles corresponding to the $j$–th education vector component) and Inequalities (11.4), then the relations

$$\lim_{t\to\infty}|Z(t)| \le \lim_{t\to\infty}\frac{4\frac{q}{(1-\theta)^2}\sum_{i=1}^m|a_i|}{t\sum_{i=1}^m a_i^2} = 0$$

are valid, therefore $\lim_{t\to\infty} Z(t) = 0$.

This proves the theorem.

Corollary 11.3. If there were some forgetful robot existing for an infinitely long period of time, then in the course of time its abilities would tend to zero, i.e. vanish.

## 12. WORK AND WILLPOWER OF EMOTIONAL ROBOTS

It is easy to see that in the univariate case (when the goal $A$ is defined by one value) the value of goal achievement by the end of the $n$-th complete education cycle satisfies the relation $\delta(t) = \dfrac{V_{l_n,i_n}^{[n]}(t)}{A}$.

Let us introduce the following definitions.

<u>Definition 12.1</u>. Education process work on achievement of the goal $A$ is the function $X(t) = \int_0^t \delta(\tau)d\tau$, where the subintegral function is a function of value of the goal $A$ achievement.

<u>Definition 12.2</u>. *Willpower* of a robot achieving the goal $A$ is the function 
$$Y(t) = \frac{\int_0^t \delta(\tau)d\tau}{t}.$$

It is easy to see that *Work* is measured in time units, while *Willpower* does not have units of measurement.

Now let us establish several simple theorems; their proofs are obvious from (11.3).

<u>Theorem 12.1</u>. In the univariate case the education process work of a forgetful robot achieving its goal satisfies the inequality $|X(t)| \leq \dfrac{2q}{|A|(1-\theta)^2} t$.

<u>Theorem 12.2</u>. In the univariate case the willpower of a forgetful robot achieving its goal satisfies the inequality $|Y(t)| \leq \dfrac{2q}{|A|(1-\theta)^2}$.

<u>Theorem 12.3</u>, In the multivariate case (the goal is a vector), the education process work of the forgetful robot achieving its goal satisfies the inequality
$$|X(t)| \leq \frac{2\dfrac{q}{(1-\theta)^2}\sum_{i=1}^{m}|a_i|}{\sum_{i=1}^{m}a_i^2} t.$$

<u>Theorem 12.4</u>, In multivariate case, the willpower of the forgetful robot achieving its goal satisfies the inequality

$$|Y(t)| \leq \frac{2\dfrac{q}{(1-\theta)^2}\sum_{i=1}^{m}|a_i|}{\sum_{i=1}^{m}a_i^2}. \tag{12.1}$$

Corollary 12.4. A forgetful robot with the unlimited willpower does not exist.

Proof. Since (12.1) holds true, the forgetful robot's willpower is limited. Hence, the corollary is proved.

Suppose the human's willpower similarly to the robot's willpower is described by Definition 12.2.

Let us introduce one more definition.

Definition 12.3. A robot is dangerous to a man when a modulus of its willpower becomes asymptotically (with time tending to infinity) more than a human willpower modulus at any time point of the man's life.

Theorem 12.3. A robot with an absolute memory and tantamount positive emotions is dangerous to a man.
Proof. Since all the memory coefficients of the robot with an absolute memory are equal to 1, then for tantamount positive emotions (considering (11.2)) the robot's education resulting from infinite quantity of education cycles is equal to infinity, i.e. satisfies the relation

$$\lim_{t \to \infty} Y(t) = \lim_{t \to \infty} \frac{\int_0^t \delta(\tau) d\tau}{t} = \infty.$$

Even having all his emotions positive, a regular human being does not have an absolute memory, his\her emotions are limited [8], so, according to Theorem 12.4 a human willpower is limited, i.e. it is less than an asymptotic willpower of a robot with an absolute memory and tantamount positive emotions.
The proof is complete.

Since a single person by nature cannot physically exist forever, his\her willpower is always finite.
After a forerunning robot rests in peace, the information from its memory can be downloaded to the successive robot's memory (together with the information on numerical values of all the previous generations of robots). This provides continuous existence of a single robot's intelligence with the time tending to infinity. As a result of such continuous existence of generations of robots and passing on positive tantamount emotions from "ancestors" to "successors" provided with an absolute memory we will surely come up to a moment when a robot becomes dangerous to a man. So, we may conclude that in order to avoid this danger for a human being it is necessary at least to design forgetful robots (robots with a non-absolute memory).
Suppose the following relation holds true:

$$\rho(\overline{\theta}_{max},t) = \max_{\overline{\theta}} |\int_0^t \delta(\overline{\theta},\tau)d\tau| \qquad (12.2)$$

with: $\overline{\theta}$ the varied collections of memory coefficients; $\overline{\theta}_{max}$ the vector of memory coefficients, under which the function $|\int_0^t \delta(\overline{\theta},\tau)d\tau|$ reaches its maximum.

From Definition 12.1 and Formula (12.2) we derive the following definition.

<u>Definition 12.4</u>. The efficiency coefficient $\mu(t)$ of the education process is a value satisfying the relation

$$\mu(t) = \frac{X(t)\, sign[\int_0^t \delta(\overline{\theta}_{max},\tau)d\tau]}{\rho(\overline{\theta}_{max},t)} =$$

$$= \frac{\int_0^t \delta(\tau)d\tau\, sign[\int_0^t \delta(\overline{\theta}_{max},\tau)d\tau]}{\max_{\overline{\theta}} |\int_0^t \delta(\overline{\theta},\tau)d\tau|}.$$

It is easy to see that the education process efficiency coefficient has no units of measurement and the condition $\mu(t) \in [-1,1]$ is valid for it. Obviously, the more is $\mu(t)$ under the given memory coefficients, the closer gets a robot to the most effective education.

Provided that $sign[X(t)]\, sign[\int_0^t \delta(\overline{\theta}_{max},\tau)d\tau] > 0$ holds true, the value $\mu(t)$ satisfies the relation $\mu(t) \in (0,1]$, which means that directions of real and effective education processes coincide.

It should be noted that the efficiency coefficient makes it possible to assess "natural" robot properties (memory coefficients) in terms of education process effectiveness.

## 13. ROBOT's TEMPERAMENT MODEL

This chapter gives mathematical interpretation of robot's temperaments.
<u>Definition 13.1</u>. The elementary temperament $w_i(t)$ is a derivative of the function of momentary emotions module $M_i(t)$ with respect to the time $t$, i.e.

$w_i(t) = \dfrac{d|M_i(t)|}{dt}$, where $i$ is the robot's number in a group, $\dfrac{d|M_i(t)|}{dt} > 0$, $i = \overline{1,n}$, $n$ is the quantity of robots in a group.

Psychological studies say that we can hardly meet a human being with the pronounced temperament of one certain type. As human beings are analogues of robots, let us give the following definition.

<u>Definition 13.2</u>. The robot's temperament $L$ is a function satisfying the relation

$$L = \dfrac{1}{an}\sum_{i=1}^{n}\left(\max_{t}\left|\dfrac{dM_i(t)}{dt}\right|\right), \quad \text{with } a = \max_{i,t}\left|\dfrac{dM_i(t)}{dt}\right|, \; i \in [1,n].$$

It is easy to see that the suggested rule allows us to find a temperament of some certain robot only relative to its (sub)group.

The results of investigations of human temperament and its numerical values [3] can obviously be applied to robots (see Table 13.1).

Table 13.1. **Variation intervals of robot's temperament values**

| Robot's temperament type | Variation intervals of temperament numerical value |
|---|---|
| melancholic | (0; 0,3) |
| phlegmatic | (0,3; 0,5) |
| sanguine | (0,5; 0,8) |
| choleric | (0,8; 1) |

Intervals given in Table 13.1. allow us to introduce a concept of temperament of a group of robots (group's temperament).

In Chapter 2 we mentioned an example which can be described by the function $M(t) = P\sin\left(\dfrac{\pi}{t^0}t\right)$ with: $P=const$; $t^0$ the time step lenght.

Similarly to this example we define a set of emotions; each of them for the robot $i$. takes a form:

$$M_i(t) = P_i \sin\left(\dfrac{\pi}{t_i^0}t\right)$$

with: $P_i = const$, $t_i^0$ the lenght of the time step $i$, $i = \overline{1,n}$.

It is easy to see that in this case the robot's temperament $L_i$ can be defined by the following formula

$$L_i = \frac{\dfrac{|P_i|}{t_i^0}}{\max\limits_{i=\overline{1,n}} \dfrac{|P_i|}{t_i^0}}.$$

<u>Definition 13.3</u>. The temperament $N$ of the group of robots is the average temperament of robots belonging to this group.

With the definition of finding $N$ in mind, we can use the following formula:

$$N = \frac{\sum\limits_{i=1}^{n} L_i}{n}. \qquad (13.1)$$

Having found $N$ by (13.1), we can define the temperament type by associating values from the right column in Table 13.1 with the left column; depending on what interval $N$ belongs to, the group of robots can be either melancholic, or phlegmatic, or sanguine, or choleric.

## 14. INVESTIGATION OF PSYCHOLOGICAL PROCESS DYNAMICS IN A GROUP OF ROBOTS

Here we consider the cases when some processes occur in a group of robots with time. Terms and conventional signs used in this Chapter are the same as in Chapter 9.

The following statements are obvious.

<u>Theorem 14.1</u>. If with the course of time $\cos(\alpha(t)) \underset{t \to t_0}{\to} 0$, then a group of robots tends to the unstable emotional situation.

<u>Theorem 14.2.</u> If with the course of time $\cos(\alpha(t)) \underset{t \to t_0}{\to} 1$, then emotional activity tends a group of robots to get serried (united).

<u>Theorem 14.3.</u> If with the course of time $\cos(\alpha(t)) \underset{t \to t_0}{\to} -1$, then emotional activity tends a group of robots to get dissociated (disunited).

<u>Corollary 14.3.1.</u> If $\cos(\alpha(t)) \underset{t \to t_0}{\to} -1$ and $\left\| \overline{|a(t)|} - \overline{|b(t)|} \right\| \underset{t \to t_0}{\to} 0$, $|a(t)|$ is large, then there is a threat of conflict in its acutest form in the group.

Note that the point $t_0$ mentioned in Theorems 14.1–14.3 and Corollary 14.3.1 is the time value corresponding to the defined events in the statement given above.

Corollary 14.3.2. If the conditions of Corollary 14.3.1 are valid for emotion and education vectors simultaneously, then there is a threat of conflict in its acutest form (i.e. simultaneous emotional and educational conflicts) in the group of robots.

Based on the things given above we introduce the following definition.

Definition 14.1. The measure of emotional or educational relationship between sub-groups of robot within a set are the unitless values $\gamma(t), \varphi(t)$ satisfying the relations $\gamma(t) = \cos\left(\overline{a(t)}, \overline{b(t)}\right), \varphi(t) = \cos\left(\overline{x(t)}, \overline{y(t)}\right)$ where $\overline{a(t)}, \overline{b(t)}$ are education vectors, and $\overline{x(t)}, \overline{y(t)}$ are emotion vectors.

So, the emotional condition of the two sub-groups altogether can be described by the vector $\overline{c} = (\gamma(t), \varphi(t))$.

It is easy to see that if the relations $\gamma(t) \in (0,1]$ or $\varphi(t) \in (0,1]$ are valid at the moment of time $t$, then there is an emotional or educational concordance correspondingly between the sub-groups; and vice versa, if $\gamma(t) \in [-1,0)$ or $\varphi(t) \in [-1,0)$, then there is an emotional or educational rivalry correspondingly between the sub-groups within the set of robots. The cases $\gamma(t) = 0$ or $\varphi(t) = 0$ are responsible for borderline situations in between educational or snap-emotional rivalry and concordance. The case corresponding to the inequality $\gamma(t)\varphi(t) < 0$ defines educational concordance or emotional rivalry and vice versa.

It is obvious that the larger are $\gamma(t)$ or $\varphi(t)$ with their values positive, the more "benevolent", i.e. concordant is the atmosphere in the set of robots; and the smaller are $\gamma(t)$ or $\varphi(t)$ with their values negative, the stronger is the rivalry between the robots. The following statement holds true: if $\gamma(t) < 0$ and $\left|\overline{a(t)}\right| > \left|\overline{b(t)}\right|$, then the sub-group which education is described by the vector $\overline{b(t)}$ may be reeducated in favour of the sub-group with the education $\overline{a(t)}$.

Let us formulate the following theorem.

Theorem 14.4. If $\cos\left(\overline{a}, \overline{b}\right) = -1$ and $\left|\overline{a}\right| = \left|\overline{b}\right|$ with $n$ equal to 2 or 3, then $\overline{a} = -\overline{b}$.

Proof. As the first statement of the theorem is valid, then in a two- and three-dimensional space the vectors $\overline{a}$ and $\overline{b}$ are collinear, i.e. $\overline{a} = k\overline{b}$. By virtue of the second statement, the coefficient $k$ satisfies the equality $k = \pm 1$, and since $\cos\left(\overline{a},\overline{b}\right) = -1$, then $k = -1$, consiquently $\overline{a} = -\overline{b}$.

Under $\overline{a} = -\overline{b}$ the relations $\cos\left(\overline{a},\overline{b}\right) = -1$ and $\left|\overline{a}\right| = \left|\overline{b}\right|$ obviously hold true.

This allows us also to formulate <u>Theorem 14.5</u> for two- and three-dimensional vectors: In order $\overline{a} = -\overline{b}$, it is necessary and sufficient that conditions $\cos\left(\overline{a},\overline{b}\right) = -1$ and $\left|\overline{a}\right| = \left|\overline{b}\right|$ hold true simultaneously.

Due to Theorem 14.5 we can generalize Theorem 14.4: under the dimensionality of education and emotion vectors less than four, for the worst confrontation within the set of robots with nonzero vectors of psychoemotional states it is necessary and sufficient that the sum vector of emotions or educations is to be equal to the vector with zero components.

In the conclusion of this chapter we should note that Theorems 14.1, 14.3 and Corollaries 14.3.1, 14.3.2 allow us to assess the tendency of the set of robots to critical emotional situations. And in case these situations are undesirable, the mentioned theorems and corollaries substantiate the necessity of effecting the robots with subjects which are able to kill this tendency.

## 15. RULES AND FORECAST OF EMOTIONAL SELECTION OF ROBOTS

Using mathematical definitions of robot's psychological characteristics considered above in this chapter we try to describe one of the algorithms of robot's emotional behavior.

Suppose a robot has got an emotional selection problem: he is supposed to decide in favor of either the first or the second player (educator) depending on his education.

Below we suggest the rules of making an emotional decision for such robots. Assume that the robot has only positive emotions. Now suppose the robot's memory coefficients $\theta_{i,j}$ satisfy the correlation $0 \leq \theta_{i,j} \leq 1$, where $i = \overline{1,\infty}$, the equality $j = 1$ meets robot's memory coefficients for the first educator, the equality $j = 2$ meets memory coefficients for the second educator.

We adduce the First rule of the alternate selection based on the emotional selection. This rule can easily be implemented in computer modeling of the robot's emotional behavior.

Suppose a robot is simultaneously effected by two players initiating robot's emotions. At the time point of stimulus (subject) $t_i$ effect the first player initiates the emotion $M_{1,i}$ causing the elementary education $R_{1,i}$ equal to $\int_0^{t_i} M_{1,i}(\tau)d\tau$, and the education $\overline{B_1} = (R_1, 0)$ where e.g. for the robot absolute memory the formula $R_1 = \sum_{k=1}^{l} \int_0^{t_k} M_{1,k}(\tau)d\tau$ holds true, and the second player initiates a zero emotion at the same time.

At the time point $t_j$ the second player initiates the emotion $M_{2,j}$ causing the elementary education $R_{2,j} = \int_0^{t_j} M_{2,j}(\tau)d\tau$ where $i \neq j$, and the education of the second player $\overline{B_2} = (0, R_2)$ where e.g. for the robot's absolute memory the formula $R_2 = \sum_{k=1}^{l} \int_0^{t_k} R_{2,k}(\tau)d\tau$ holds true, and the first player initiates a zero emotion at the same time.

Let us introduce the general education vector $\overline{V}$ equal to $(R_1, R_2)$ where vector components are sum educations obtained in the time $t$ of effects of the first and second player subjects where $t = \sum_{k=1}^{l} t_k$, and $l$ is a total number of emotional effects of both players' subjects upon the robot.

With these designations introduced, the rule of deciding in favor of the first or the second player can be formulated as follows: if the angle between $\overline{V}$ and $\overline{B_1}$ is less than the angle between $\overline{V}$ and $\overline{B_2}$ then the robot decides in favor of the first player; if the first angle is wider than the second one, it means that the decision is made in favor of the second player; but if the angles are equal the selection is not supposed to be performed.

It is not very difficult to apply the first rule described above in case there are more than two players. For example, if we want to implement this rule for modeling emotional behavior of a robot it is enough to enter the number of momentary sum educations equal to the number of players, and the number of components of the general education vector is to be also increased. The minimal angle between the general education vector and the education vector of each player defines the alternate selection in favor of this or that player.

Note that the first rule is valid not only for scalar values of sum educations and emotions, but also for the cases when they have a form of vector.

Definition 15.1. Critical angle of alternate selection is an angle (between the education vector and the general education vector) defining ambiguity of a robot while making a decision in favor of the first or the second player.

Now we adduce the Second rule of alternate selection based on comparing moduli of vectors of the sum educations $\overline{R}_1$ and $\overline{R}_2$. This rule can be re-formulated as follows: if $|\overline{R}_1| > |\overline{R}_2|$ holds true, then the decision is made in favor of the first player; if $|\overline{R}_1| < |\overline{R}_2|$ holds true, then the decision is made in favor of the second player; if $|\overline{R}_1| = |\overline{R}_2|$, then the decision is not made.

Theorem 15.1. The First and the Second rules of alternate selection are equivalent to each other.

Proof. Suppose $\alpha$ is the angle between $\overline{B}_1$ and $\overline{V}$, and $\beta$ is the angle between $\overline{B}_2$ and $\overline{V}$. Then according to vector algebra rules, the relations

$$\cos\alpha = \frac{\sqrt{\sum_{i=1}^{2n} B_{1,i}^2}}{\sqrt{\sum_{i=1}^{n} R_{1,i}^2 + \sum_{i=n+1}^{2n} R_{2,i}^2}} = \frac{\sqrt{\sum_{i=1}^{n} R_{1,i}^2}}{\sqrt{\sum_{i=1}^{n} R_{1,i}^2 + \sum_{i=n+1}^{2n} R_{2,i}^2}} = \frac{|\overline{R}_1|}{|\overline{V}|}, \quad (15.1)$$

$$\cos\beta = \frac{\sqrt{\sum_{i=1}^{2n} B_{2,i}^2}}{\sqrt{\sum_{i=1}^{n} R_{1,i}^2 + \sum_{i=n+1}^{2n} R_{2,i}^2}} = \frac{\sqrt{\sum_{i=1}^{n} R_{2,i}^2}}{\sqrt{\sum_{i=1}^{n} R_{1,i}^2 + \sum_{i=n+1}^{2n} R_{2,i}^2}} = \frac{|\overline{R}_2|}{|\overline{V}|} \quad (15.2)$$

hold true.

Obviously, if $\alpha > \beta$, $0 < \alpha < \frac{\pi}{2}, 0 < \beta < \frac{\pi}{2}$, then $|\overline{R}_1| < |\overline{R}_2|$; if $\alpha < \beta$, $0 < \alpha < \frac{\pi}{2}, 0 < \beta < \frac{\pi}{2}$, then $|\overline{R}_1| > |\overline{R}_2|$; if $\alpha = \beta$, then $|\overline{R}_1| = |\overline{R}_2|$. So, we proved that the First rule implies the Second one to be valid.

Let us prove that the Second rule implies the First one.

Suppose $\left|\overline{R_1}\right| < \left|\overline{R_2}\right|$ holds true. Then by virtue of (15.1) and (15.2) the inequality $\alpha > \beta$ is inevitable under $0 < \alpha < \dfrac{\pi}{2}$, $0 < \beta < \dfrac{\pi}{2}$.

Validity of the following statements is proved similarly: if $\left|\overline{R_1}\right| > \left|\overline{R_2}\right|$, then $\alpha < \beta$ under $0 < \alpha < \dfrac{\pi}{2}$, $0 < \beta < \dfrac{\pi}{2}$; if $\left|\overline{R_1}\right| = \left|\overline{R_2}\right|$, then $\alpha = \beta$.

The proof is complete.

<u>Theorem 15.2</u>. If two vectors do not have common nonzero coordinates, then these vectors are orthogonal.

<u>Proof</u>. Since according to the theorem statement the vectors do not have common nonzero coordinates, then without breaking the integrity these vectors can take a form $\overline{a} = (a_1, a_2, ..., a_n, 0, 0, ..., 0)$, $\overline{b} = (0, 0, ..., 0, b_{n+1}, b_{n+1}, ..., b_m)$.

It is obvious that the scalar product of $\overline{a}$ and $\overline{b}$ equals to zero. It means that the vectors are orthogonal. This proves the theorem.

<u>Corollary 15.2</u>. The vectors $\overline{B_1}$ and $\overline{B_2}$ are orthogonal.

<u>Proof</u>. Since, according to the designations of the First rule, $\overline{B_1}$ and $\overline{B_2}$ do not have nonzero common coordinates, then by virtue of Theorem 15.2 these vectors are orthogonal.

Let us prove one of the properties of alternate emotional selection.

<u>Theorem 15.3</u>. The alternate selection critical angle is equal to $\dfrac{\pi}{4}$.

<u>Proof</u>. Let us note that $\overline{V} = \overline{B_1} + \overline{B_2}$ is valid. According to the parallelogram law for composition of two vectors, $\overline{V}$ is a diagonal of the parallelogram with the adjacent sides $\overline{B_1}$ and $\overline{B_2}$. By virtue of Corollary 15.2 these sides are orthogonal, so $\alpha + \beta = \dfrac{\pi}{2}$ is valid. According to Definition 15.1 and the First rule of alternate selection, $\alpha = \beta$ holds true, i.e. the alternate selection critical angle is equal to $\dfrac{\pi}{4}$.

<u>Definition 15.2</u>. *Stupor* is a state of ambiguity or uncertainty of a robot making emotional selection.

Assume that effects of the first and the second players upon the robot correspond to tantamount emotions yielding the elementary education $R_0$. Suppose robot memory coefficients corresponding to emotions resulted from the first player effect are constant and equal to $\theta_1$, and coefficients corresponding to emotional effects of the second player are equal to $\theta_2$. Also assume that $\theta_i \in [0,1)$, $i \in \{1, 2\}$, and robot's emotional memory of the first player effect is fully kept while the second player is making his effect and vice versa.

Then, according to Chapter 3 and the Second rule of alternate selection the following equality is obvious:

$$R_0 \frac{1-\theta_1^j}{1-\theta_1} = R_0 \frac{1-\theta_2^q}{1-\theta_2}, \qquad (15.3)$$

with $j$, $q$ the quantity of emotional effects upon the robot (emotions are initiated by the first and the second player correspondingly).

Eq. (15.3) is equivalent to the relation

$$\frac{1-\theta_1^j}{1-\theta_1} = \frac{1-\theta_2^q}{1-\theta_2}. \qquad (15.4)$$

It is easy to see that under the assumptions mentioned above Eq. (15.4) defines the necessary and sufficient condition for the stupor initiated by a single-type emotion. This condition can be easily generalized in case we need to consider emotions and an education defined by vectors (in this connection it is necessary to consider various pairs of coefficients $\theta_{1,k}$ и $\theta_{2,k}$, where $k$ indicates the order number of an emotion in the robot's emotion vector).

<u>Theorem 15.4</u> is obvious. If a robot has only tantamount emotions and constant memory coefficients corresponding to each of the two players, and Eq. (15.4) is valid for each of educations, then the robot is stuporous (in stupor) with respect to all its emotions.

Obviously, the robot never comes to this state of stupor if with any $j$ and $q$ ($j > 1$, $q > 1$) and given $\theta_1$ and $\theta_2$ Eq. (15.4) is not valid.

Let us introduce one more definition.

<u>Definition 15.4</u>. Anti-stupor coefficients are the memory coefficients $\theta_1$ and $\theta_2$ for which under any integral values $j$ and $q$ ($j > 1$, $q > 1$) Eq. (15.4) does not become valid.

<u>Theorem 15.5.</u> Anti-stupor memory coefficients exist.
<u>Proof.</u> Let us show that there exist the memory coefficients $\theta_1$ and $\theta_2$ which are not the roots of Eq. (15.4) under any integral values $j$ and $q$ ($j > 1$, $q > 1$).

Obviously, Eq. (15.4) is equivalent to
$$\theta_2^j(1-\theta_1) - \theta_2(1-\theta_1^q) + (\theta_1 - \theta_1^q) = 0. \tag{15.5}$$

Suppose the following equalities
$$\theta_1 = \frac{1}{2}, \quad \theta_2 = \frac{1}{3} \tag{15.6}$$
hold true.

If we substitute Eqs. (15.6) into Eq. (15.5) and make transformations, as a result we get
$$3^j(2^{q-1} - 1) + 3^{j-1}(1 - 2^q) + 2^{q-1} = 0. \tag{15.7}$$

Considering that $y = 3^{j-1}$, Eq. (15.7) takes the form
$$3y(2^{q-1} - 1) + y(1 - 2^q) + 2^{q-1} = 0. \tag{15.8}$$

Solving (15.8) relative to $y$ we get the formula $y = -\dfrac{2^{q-1}}{2^{q-1} - 2}$ equivalent to the relation
$$3^{j-1} = -\frac{2^{q-1}}{2^{q-1} - 2}. \tag{15.9}$$

Since according to the theorem statement $j > 1$ is valid, then for any $j$ and any $q > 2$ the positive value in the left part of Eq. (15.9) is equal to the negative value in the right part of Eq. (15.9). So, we get the contradiction. Consequently, $\theta_1 = \frac{1}{2}, \theta_2 = \frac{1}{3}$ are not the roots of Eq. (15.4) with any values $j > 1$ and $q > 2$.

Now let us consider the case when $q=2$.

It easy to see that Eq. (15.8) in this case takes the form $2 = 0$, i.e. under the memory coefficients $\theta_1 = \frac{1}{2}, \theta_2 = \frac{1}{3}$ this equation has no solution.

So, with any $j > 1, q > 1$, there are such memory coefficient values under which Eq. (15.4) makes no sense. Consequently, anti-stupor memory coefficients do exist.

This completes the proof of Theorem 15.5.

<u>Corollary 15.5</u>. For two players the coefficients $\theta_1 = \frac{1}{2}, \theta_2 = \frac{1}{3}$ are anti-stupor memory coefficients.

Its <u>proof</u> is evident directly from argumentations given in the proof of Theorem 15.5.

Eq. (15.4) and Corollary 15.5 allow us to forecast the robot's behavior and see whether our robot may get into emotional stupor.

Reasoning from the things said above we can state that the 'resolute' or 'purposeful' robot is a machine for which an alternate selection angle never equals $\frac{\pi}{4}$, or Eq. (15.4) never holds true, or its memory coefficients are anti-stuporous, so that this machine does not get stuporous regarding all the components of the education vector.

## 16. GENERALIZATION OF ROBOT'S EMOTIONAL BEHAVIOR RULES IN CASE THE NUMBER OF PLAYERS INTERACTING WITH THE ROBOT IS ARBITRARY (NOT SPECIFIED)

### 16.1. FIRST RULE OF ALTERNATE SELECTION

Assume a robot is effected by $n$ players nonsimultaneously. Suppose they initiate only positive emotions and the robot performs an absolute emotional memory i.e. its memory coefficients $\theta_{i,j}$ satisfy the identity $\theta_{i,j} \equiv 1$, where $i = \overline{1, m_j}$, $j = \overline{1, n}$. Correspondingly, $m_j$ is the quantity of subject effects of the $j$-th player.

At the time point $t_{1,k}$ (with $k = \overline{1, m_1}$) the first player initiates the emotion $M_{1,k}$ causing the elementary education $R_{1,k} = \int_0^{t_{1,k}} M_{1,k}(\tau)d\tau$ and education $\overline{B_1} = (R_1, \underbrace{0,...,0}_{n-1})$ with $R_1 = \sum_{l=1}^{m_1} \int_0^{t_{1,l}} M_{1,l}(\tau)d\tau$. At the same time all the rest $n-1$ players initiate zero emotions.

At $t_{i,k}$ where $k = \overline{1, m_i}$, $t_{i,k} > t_{i_1,k_1}$, with $i > i_1$ and $k_1 = \overline{1, m_{i_1}}$ the player $i$ initiates the emotion $M_{i,k}$ causing the elementary education $R_{i,k} = \int_0^{t_{i,k}} M_{i,k}(\tau)d\tau$ and education $\overline{B_i} = (0,...,0, \underbrace{R_i}_{i-\text{йёåìåíò}}, 0,...,0)$ where $R_i = \sum_{l=1}^{m_i} \int_0^{t_{i,l}} M_{i,l}(\tau)d\tau$. At the same time all the rest $n-1$ players initiate zero emotions.

Let us introduce the general education vector $\overline{V} = (R_1, R_2,..., R_n)$ where components are sum educations (resulting from all the players' subjects) obtained in the full course of the effect time $t$, with $t = \sum_{l=1}^{n}\sum_{k=1}^{m_l} t_{l,k}$.

With these designations introduced, the rule of deciding in favor of this or that player can be formulated as follows: the emotional decision is made in favor of a player for which $\min \angle(\overline{V}, \overline{B}_i)$ is reached with $i = \overline{1, n}$ (this emotional decision is made in favor of the player $i$). In case the minimum is reached under several $i$ simultaneously, the emotional selection is not supposed to be performed and the decision is not made.

The given rule can be generalized in case the player's effect initiates not just a single emotion, but a full vector of emotions. Thereby at $t_{i,k}$ with $k = \overline{1, m_i}$, $t_{i,k} > t_{i_1, k_1}$ where $i > i_1$ and $k_1 = \overline{1, m_{i_1}}$ the player $i$ initiates the robot's emotion vector $\overline{M}_{1,k} = (M^1_{i,k}, ..., M^r_{i,k})$ which entails the vector of elementary educations

$$\overline{R}_{i,k} = (R^1_{i,k}, ..., R^r_{i,k}), R^j_{i,k} = \int_0^{t_{i,k}} M^j_{i,k}(\tau) d\tau$$ and the education

$$\overline{B}_i = (0, ..., 0, \underbrace{\overline{R}_i}_{i-\text{ýëåìåíò}}, 0, ..., 0) = (0, ..., 0, R^1_i, ..., R^r_i, 0, ..., 0)$$ with $R^j_i = \sum_{l=1}^{m_i} \int_0^{t_{i,l}} M^j_{i,l}(\tau) d\tau$. At the same time all the rest players $n - 1$ initiate zero emotions.

In this case the general education vector takes the form: $\overline{V} = (\overline{R}_1, \overline{R}_2, ..., \overline{R}_n) = (R^1_1, ..., R^r_1, ..., R^1_n, ..., R^r_n)$

Further reasoning are quite the same as those ones given above concerning the case when the player's effect initiates one robot's emotion.

### 16.2. SECOND RULE OF ALTERNATE SELECTION

The Second rule of alternate selection is based on comparison of moduli of vectors of the sum educations $\overline{B}_i$ with $i = \overline{1, n}$. This rule can be re-formulated as follows: the emotional decision is made in favor of a player for which $\max |\overline{B}_i|$ is reached with $i = \overline{1, n}$ (this emotional decision is made in favor of the player $i$). In case the maximum length is reached under several values of $i$ simultaneously, the emotional selection is not supposed to be performed and the decision is not made.

### 16.3. ORTHOGONALITY OF EDUCATION VECTORS AND EQUIVALENCE OF ALTERNATE SELECTION RULES

As it was mentioned above, in this paper we use Cartesian rectangular coordinates. According to Theorem 15.2 two vectors which do not have common nonzero coordinates are orthogonal.

Thus each pair of vectors $\overline{B}_1, ..., \overline{B}_n$ is orthogonal.

<u>Theorem 16.1.</u> The First and Second rules of alternate selection are equivalent to each other.

Proof. Let $\alpha_i = \angle(\overline{V}, \overline{B}_i)$, $0 < \alpha_i < \dfrac{\pi}{2}$, $i = \overline{1,n}$ is the angle between $\overline{V}$ and $\overline{B}_i$. According to the rules of vector algebra and orthogonality of $\overline{B}_1, ..., \overline{B}_n$ the following relation holds true: $\cos \alpha_i = \dfrac{|\overline{B}_i|}{|\overline{V}|}$.

Obviously, if $\min \angle(\overline{V}, \overline{B}_i)$ is reached under $i = k$, then according to the First rule of alternate selection the decision is made in favor of the player $k$. At that from the formula given above it follows that $|\overline{B}_k| > |\overline{B}_j|$ with $j \neq k$. Thus $\max |\overline{B}_i|$ is reached under $i = k$. The last one describes the Second rule of alternate selection.

On the other hand if $\max |\overline{B}_i|$ is reached under $i = k$ then according to the Second rule of alternate selection the decision is made in favor of the player $k$. At that $|\overline{B}_k| > |\overline{B}_j|$ with $j \neq k$ holds true, and following the formula given above we get $\alpha_k < \alpha_j$ with $j \neq k$. From the things stated above we conclude that $\min \angle(\overline{V}, \overline{B}_i)$ is reached under $i = k$, and this is according to the First rule of alternate selection.

This completes the proof.

## 17. EMOTIONAL SELECTION AND CONFLICTS BETWEEN ROBOTS

It is not difficult to see that Eq. (15.4) coincides completely with Formula (3.7) obtained while describing a conflict between two robots with equal tantamount emotions. This fact makes us conclude that inner emotional conflicts of a robot can be described by the same formulas as conflicts between different robots, and consequently, theories applicable for groups of robots can be successfully used for inner emotional conflicts of a single robot without any alterations.

As an example of this we present the following theorem.

Theorem 17.1. If two uniformly forgetful robots have the same (equal) tantamount emotions, then there are such robot memory coefficients that the robots never get into education conflict.

Proof. For conflicting robots Eq. (3.7) holds true; if tantamount emotions are equal (3.7) is transformed into Eq. (15.4). According to Theorem 15.5 there exist anti-stupor coefficients transforming Eq. (15.4) to a strict inequality. But at the same time these anti-stupor coefficients are the memory coefficients of two different robots, and moreover, with these coefficients robots would never get into conflict.

This completes the proof.

Logic makes us introduce a new definition.

Definition 17.1. Anti-conflict memory coefficients are memory coefficients of two different robots under which the robots never get into conflict.

Now it is time to give the following theorem.

Theorem 17.2. Anti-conflict memory coefficients of two uniformly forgetful robots with equal tantamount emotions coincide with anti-stupor coefficients.

Proof is analogous to the one of Theorem 17.1.

Corollary 17.2. When the conditions of Theorem 17.2 are valid, then the memory coefficients of two robots $\theta_1 = \frac{1}{2}$ and $\theta_2 = \frac{1}{3}$ are anti-conflict.

Proof. According to Corollary 15.5, anti-stupor coefficients satisfy the equalities $\theta_1 = \frac{1}{2}$, $\theta_2 = \frac{1}{3}$. By virtue of Theorem 17.2, anti-stupor coefficients are anti-conflict ones.

So, the corollary is proved.

## 18. DIAGNOSTICS OF EMOTIONAL ROBOT's "MENTAL DISEASES"

Let us recall the definition of robot's emotion given in the beginning of the book for better understanding of this chapter.

Definition 1.3. The robot's inner emotional experience function $M(t)$ is called an 'emotion' if it satisfies the following conditions:

1. Function domain of $M(t)$: $t \in [0, t^0]$, $t^0 > 0$;

2. $t^0 \leq t^*$ (note that this condition is equivalent to emotion termination in case the subject effect is either over or not over yet);

3. $M(t)$ is the single-valued function;

4. $M(0) = 0$;

5. $M(t^0) = 0$;

6. $M(t)$ is the constant-sign function;

7. There is the derivative $\frac{d|M(t)|}{dt}$ within the function domain;

8. There is the only point $z$ within the function domain, such that $z \neq 0$, $z \neq t^0$ and $\left.\frac{d|M(t)|}{dt}\right|_{t=z} = 0$;

9. $\frac{d|M(t)|}{dt} > 0$ with $t < z$;

10. $\dfrac{d|M(t)|}{dt} < 0$ with $t < z$.

Let us introduce a couple more definitions.
<u>Definition 18.1</u>. Let us consider a robot to be "healthy" if its inner emotional experience function is an emotion.

<u>Definition 18.2</u>. Let us consider a robot to be "ill" if its inner emotional experience function does not satisfy at least one of the conditions in the definition of emotion.

This definition allows us to introduce such concept as seriousness or severity of a robot's disease.
Since Definition 1.3 includes 10 conditions defining a disease, then the degree of severity of this disease is characterized by $H$ taking on integral values from 1 till 10 to indicate a number of conditions which do NOT hold true (as those are the conditions under which the inner emotional experience function becomes an emotion).
The more severe is the disease the greater is $H$.

<u>Definition 18.3</u>. The vector $X$ of disease symptoms is a vector with the numbers of emotion conditions (given in Definition 1.3) which do not hold true.

<u>Definition 18.4</u>. A robot's disease with the symptom vector $X_1$ is a special case of a robot's disease with the symptom vector $X_2$ if all the elements of the symptom vector $X_2$ occur among the elements of the symptom vector $X_1$.

Below we give examples of robots' diseases.
1. Let us take some inner emotional experience function *f(t)* satisfying all the conditions of becoming an emotion except Condition #2, i.e. the function differs from an emotion and this is described by the relation $t^0 > t^*$. Obviously, in this case the disease severity degree is equal to 1. We consider that a robot having such an emotional experience function is *neurasthenic*. It is also obvious that for neurasthenia the disease symptom vector has the form $X=(2)$.

2. Let us take some inner emotional experience function *f(t)* satisfying all the conditions of becoming an emotion except Conditions #2, 5, 8, 10. $f(t) = -t^2$ is a good example of such a function. Obviously, in this case the disease severity degree is equal to 4. A robot which emotional experience function differs from an emotion regarding Conditions #2, 5, 8, 10 is *psychopathic*. For psychopathy the disease symptom vector has the form $X=(2, 5, 8, 10)$.

The forms of vectors in these examples make us conclude that symptoms of neurasthenia and psychopathy have one thing in common, which is Condition #2 unsatisfied, and, according to Definition 18.4 psychopathy is a special case of neurasthenia.

Sometimes one unsatisfied condition of the emotion definition implies that some other conditions get unsatisfied, too.

Let us consider the inner emotional experience function which has the form:

$$f(t) = P \sin\left(\frac{\pi}{t^0} t\right) - \frac{1}{2} P, \quad P = const \; P > 0, \quad t \in [0, t^0]. \qquad (18.1)$$

At first sight Function (18.1) does not satisfy only Condition #5, and the disease severity degree is equal to 1 and the symptom vector containing only one element has the form $X=(6)$. But it is not correct. Applying mathematical analysis we can conclude that if Condition #6 is unsatisfied it implies that Conditions #4, 5, 7, 9, 10 are not satisfied for the function $f(t)$, as well. I.e. the disease severity degree is equal to 6, and the symptom vector satisfies the relation $X=(4, 5, 6, 7, 9, 10)$.

The example illustrating Formula (18.1) demonstrates the method (based on mathematical analysis) for detection of the major symptom of an emotional robot disease. Elimination of this symptom directly implies that all the rest conditions become valid and satisfied.

Thus for Function (18.1) the major reason of a rather severe disease is that Condition #6 remains unsatisfied.

## 19. MODELS OF ROBOT's AMBIVALENT EMOTIONS

Suppose we have the robot's emotion vector $\overline{M}(\tau)$ defining ambivalent emotions. This vector takes the form

$$\overline{M}^j(\tau) = \left(M_i^j(\tau), \ldots, M_n^j(\tau)\right)$$

with: $n$ the quantity of displayed emotions in the robot's ambivalent emotion, $\tau$ the current time of the emotion effect.

If the education goal is known and it is defined by $\overline{A} = (A_1, \ldots, A_n)$ where $A_i = const, \; i = \overline{1, n}$ then the value of goal achievement extent $\delta$ of the education process is specified by the following equality:

$$\delta(t) = \frac{\sum_{i=1}^{n} A_i R_i^j(t)}{\sum_{i=1}^{n} A_i^2}, \qquad (19.1)$$

with: $R_i^j(t)$ the robot's education obtained as a result of effect of the $i$-th emotion (at that $R_i^j(t) = r_i^j(t) + \theta_i^j(t) R_i^{j-1}(t)$), $\theta_i^j(t)$ the memory coefficient satisfying the relation $\theta_i^j(t) \in [0, 1]$, $j$ the order number of an education time step, $t$ the time of the education process, $r_i^j(t)$ the elementary education satisfying the relation

$$r_i^j(\tau) = \int_0^\tau M_i^j(\zeta)d\zeta \ , \ t = t_{i-1} + \tau.$$

Differentiating (19.1) with respect to $t$, we obtain

$$\frac{d\delta(t)}{dt} = \frac{\sum_{i=1}^n A_i \dfrac{dR_i^j(t)}{dt}}{\sum_{i=1}^n A_i^2}. \tag{19.2}$$

According to Chapter 2 the sum emotion $V_i^j(t)$ satisfies the relation

$$V_i^j(t) = \frac{dR_i^j(t)}{dt} = \frac{dr_i^j(t)}{dt} + R_i^{j-1}(t)\frac{d\theta_i^j(t)}{dt} + \frac{dR_i^{j-1}(t)}{dt}\theta_i^j(t). \tag{19.3}$$

It easy to see that for the robot with an absolute memory this formula is equivalent to

$$V_i^j(t) = \frac{dR_i^j(t)}{dt} = M_i^j(t).$$

So, Eq. (19.2) takes the form

$$\frac{d\delta(t)}{dt} = \frac{\sum_{i=1}^n A_i V_i^j(t)}{\sum_{i=1}^n A_i^2}. \tag{19.4}$$

Modern psychologists believe that an emotion is positive if it makes an entity (a person or a robot) to approach its preset goal. Thus if $\dfrac{d\delta(t)}{dt} > 0$ holds true then the ambivalent vectorial emotion is positive; if $\dfrac{d\delta(t)}{dt} < 0$ holds true then this ambivalent emotion is negative; if $\dfrac{d\delta(t)}{dt} = 0$ holds true then it has no sign.

But modern vector algebra in the general case does not operate with such terms as "positive" or "negative" vectors. Therefore let us advanced a hypothesis that there is a unified characteristic for ambivalent emotions of the vector which specifies a sign of the ambivalent vectorial emotion. Obviously, this characteristic is a sign of the value $\dfrac{d\delta(t)}{dt}$.

Let us introduce series of definitions.

Definition 19.1. The average function $[f(t)]$ of robot's inner emotional experience is the function of the form

$$[f(t)] = \frac{\sum_{i=1}^{n} A_i V_i^j(t)}{\sum_{i=1}^{n} A_i} \qquad (19.5)$$

under the stipulation that $\sum_{i=1}^{n} A_i \neq 0$, $t \in [0, t_0]$, $t_0$ is the minimum value of all the time steps of component emotions of the ambivalent emotion vector.

Thus the average function of robot's inner emotional experience represents a special function, such that when this function is substituted for all the sum component emotions in the ambivalent emotion vector we get the value $\frac{d\delta(t)}{dt}$ equal to the value of the function without this substitution. I.e.

$$\frac{\sum_{i=1}^{n} A_i V_i^j(t)}{\sum_{i=1}^{n} A_i^2} = \frac{\sum_{i=1}^{n} A_i [f(t)]}{\sum_{i=1}^{n} A_i^2}$$

holds true for this substitution

Definition 19.2. The average emotion $[M(t)]$ is an average function of inner emotional experience which appears to be an emotion.

Definition 19.3. If an average function of inner emotional experience is not an emotion then a robot is considered to be mentally ill and an ambivalent emotion causes the disease.

Definition 19.4. The average elementary education $[D]$ is a value satisfying the relation $[D] = \int_{0}^{t_0} [M(\tau)] d\tau$.

Definition 19.5. The average education $[R]$ is a value specified by the formula

$[R] = \frac{\sum_{i=1}^{n} A_i R_i^j}{\sum_{i=1}^{n} A_i}$ with $\sum_{i=1}^{n} A_i \neq 0$.

Definition 19.6. The prevailing emotion $M_k(t)$ in the ambivalent emotion vector is an emotion for which its order number $k$ in the vector of ambivalent emotions implies that

$$\left| r_k^j - [D] \right| = \min_{i=\overline{1,n}} \left| r_i^j - [D] \right|$$

is satisfied.

Definition 19.7. The prevailing elementary education is the elementary education corresponding to the prevailing emotion.

Obviously for each current time step $j$ of the robot's education there can be its average function of inner emotional experience, average emotion, prevailing emotion, prevailing elementary education, average elementary education, average education and value characterizing an ambivalent emotion sign.

Let the emotion vector $M_i^j(t), i = \overline{1,n}$ of the vector of ambivalent emotions have the form

$$M_i^j(t) = P_i^j \sin\left(\frac{\pi}{t_0} t\right), \quad P_i^j = const \neq 0, \quad i = \overline{1,n}, \quad t \in [0, t_0]. \quad (19.6)$$

Now let us prove the theorem.

Theorem 19.1. If $\sum_{i=1}^{n} A_i P_i^j \neq 0$ then for the robot with an absolute memory the average function of inner emotional experience satisfying Eqs (19.6) is an emotion.

Proof. It is quite easy to see that with (19.6) valid the value $[f(t)]$ satisfies the following relation:

$$[f(t)] = \frac{\sum_{i=1}^{n} A_i P_i^j}{\sum_{i=1}^{n} A_i} \sin\left(\frac{\pi}{t_o} t\right). \quad (19.7)$$

Obviously (19.7) satisfies the definition of *emotion*.
Quod erat demonstrandum.

Theorem 19.2. If $\sum_{i=1}^{n} A_i > 0$ then the sign of the average emotion coincides with the sign of the ambivalent emotion
Its Proof becomes obvious when we compare Formulas (19.4) и (19.5).

Theorem 19.3. If $\sum_{i=1}^{n} A_i < 0$ then the sign of the average emotion and the sign of the ambivalent emotion are opposite.

Its Proof is analogous to the proof of Theorem 19.2.

## 20. ABSOLUTE MEMORY OF ROBOTS

Let us consider robots with memory coefficients satisfying the eqs. $\theta_i \equiv 1, \quad i = \overline{1, n}$, where $n$ is the number of time steps in the education process.

Obviously, in this case the education $R_n$ is defined by the formula

$$R_n = \sum_{i=1}^{n} r_i, \qquad (20.1)$$

where $r_i$ is the elementary education corresponding to the $i$-th time step.

According to Eq. (20.1), the infinite education process $R$ can be described by the equality

$$R = \lim_{n \to \infty} R_n = \sum_{i=1}^{\infty} r_i. \qquad (20.2)$$

Now let us formulate the following theorems.

Theorem 20.1. An infinite education process based on tantamount emotions for the robot with an absolute memory diverges.

Proof. Since the emotions are tantamount, then the equalities. $r_i = q, \quad i = \overline{1, \infty}$ are valid. By virtue of them Relation (20.2) takes the form $R = \lim_{n \to \infty} \sum_{i=1}^{n} q = q \lim_{n \to \infty} n = \pm \infty$.

The theorem is proved.

Theorem 20.2. If an infinite education process converges, then elementary educations which this process is based on tend to zero with an infinite increase in the number of time steps.

Proof. Since the education process converges, then the inequality $|\sum_{i=1}^{\infty} r_i| < \infty$ holds true. Consequently, $\lim_{i \to \infty} r_i = 0$.

The theorem is proved.

Note one more thing: the education process convergence corresponds to the education satiety presence under an increase in the number of time steps. Taking this into account we can rephrase Theorem 20.2 as follows: if an education process is satiated, then the elementary education in the basis of this process tends to zero with an infinite increase in the number of time steps.

In Chapter 1 we gave an example of an emotion which can be described by the function $M(t) = P \sin\left(\dfrac{\pi}{t^0} t\right)$, where $P = const$, $t^0$ is the time step length.

Similarly to this example, we define emotions corresponding to the i-th time step by

$$M_i(t) = P_i \sin\left(\dfrac{\pi}{t_i^0} t\right), \qquad (20.3)$$

where $P_i = const$, $t_i^0$ is the length of the i-th time step, $i = \overline{1, \infty}$.

It is easy to see that the elementary education $r_i$ corresponding to Emotion (20.3) satisfies the equality $r_i = \dfrac{2}{\pi} P_i t_i^0$.

So, by virtue of Theorem 20.2, if an education converges, the equality $\lim\limits_{i \to \infty} P_i t_i^0 = 0$ has to be valid, and this is the necessary convergence condition.

Let us prove the following theorems.

<u>Theorem 20.3</u>. If $\lim\limits_{i \to \infty} t_i^0 = 0$, then $\lim\limits_{i \to \infty} r_i = 0$.

<u>Proof</u>. According to Definition 1.3, $|P_i| < L < \infty$ is valid. Consequently, the chain of relations $\left|\lim\limits_{i \to \infty} r_i\right| = \left|\lim\limits_{i \to \infty} P_i t_i^0\right| \leq L \lim\limits_{i \to \infty} t_i^0 = 0$ holds true. This completes the proof of Theorem 18.3.

<u>Theorem 20.4</u>. If $\lim\limits_{i \to \infty} P_i = 0$, then $\lim\limits_{i \to \infty} r_i = 0$.

<u>Proof</u>. According to Definition 1.3, $\left|t_i^0\right| < S < \infty$ is valid. Consequently, the chain of relations $\left|\lim\limits_{i \to \infty} r_i\right| = \left|\lim\limits_{i \to \infty} P_i t_i^0\right| \leq S \left|\lim\limits_{i \to \infty} P_i\right| = 0$ holds true. This completes the proof of Theorem 20.4.

As is obvious from the foregoing, the condition necessary for education convergence is satisfied if $\lim\limits_{i \to \infty} t_i^0 = 0$, or $\lim\limits_{i \to \infty} P_i = 0$, or $\lim\limits_{i \to \infty} P_i t_i^0 = 0$ holds true.

The following statement is obvious as well: if there are limits of $t_i^0$ and $P_i$ under an infinite increase in the number of time steps., and $\lim_{i \to \infty} t_i^0 \neq 0$ and $\lim_{i \to \infty} P_i \neq 0$ hold true, then the education process is divergent.

The theorems proved above direct us to one of the way of designing robots with an absolute memory and without education satiety. E.g., in order to develop this kind of robots it is enough just to select the sequences of amplitudes $P_i$ and the time steps $t_i^0$ such that their limits under an infinite increase in the number of time steps $i$ are nonzero. According to Theorem 20.1, an example of a divergent education is the education with tantamount emotions, i.e. when the conditions
$P_i = P = const, \quad t_i^0 = t^0 = const, \quad i = \overline{1, \infty}$ hold true.

To build a robot with a satiated education one may select the predetermined convergent series as the infinite education, then on its basis define the sequences $P_i$, $t_i^0$, $i = \overline{1, \infty}$ satisfying Definition 1.3 and the statements of Theorems 20.3 or 20.4, and then based on this selection preset the emotions for each of time steps by Formula (20.3).

Based on Chapter 3 we can state that an education conflict between two robots with an absolute memory by the time point $t$ occurs if the following conditions are satisfied:

$$\sum_{k=1}^{i} r_k^{[1]} = \sum_{k=1}^{j} r_k^{[2]}, \quad t = \sum_{k=1}^{i} \tau_k^{[1]} = \sum_{k=1}^{j} \tau_k^{[2]}, \quad (20.4)$$

where $r_k^{[1]}$, $r_k^{[2]}$ are the elementary educations of the first and the second robot, $\tau_k^{[1]}$, $\tau_k^{[2]}$ are the corresponding education time steps of these robots.

Let the robots get their educations based on tantamount emotions with the corresponding elementary educations $r_0^{[1]}$ and $r_0^{[2]}$. Consequently, reasoning from conflict relations (20.4) we obtain $ir_0^{[1]} = jr_0^{[2]}$, i.e. conditions for two robots to get into conflict at the time point $t$ takes the form

$$\frac{i}{j} = \frac{r_0^{[2]}}{r_0^{[1]}}, \quad t = \sum_{k=1}^{i} \tau_k^{[1]} = \sum_{k=1}^{j} \tau_k^{[2]}. \quad (20.5)$$

If
$$\tau_k^{[1]} = \tau^{[1]} = const, \quad \tau_k^{[2]} = \tau^{[2]} = const, \quad (20.6)$$
hold true, then Relations (20.5) are equivalent to the formula

$\dfrac{i}{j} = \dfrac{r_0^{[2]}}{r_0^{[1]}} = \dfrac{\tau^{[2]}}{\tau^{[1]}}$, which defines the conditions for an education conflict to start between two robots with an absolute memory under tantamount emotions for each robot and equal time steps of these emotions.

For emotions given by Formula (20.3) and considering Eqs. (20.6) we get the following equality:

$$\frac{i}{j} = \frac{P^{[2]} t_{[2]}^0}{P^{[1]} t_{[1]}^0} = \frac{t_{[2]}^0}{t_{[1]}^0}, \qquad (20.7)$$

where $P^{[1]}, P^{[2]}, t_{[1]}^0, t_{[2]}^0$ are the amplitudes of emotions and the values of time steps of the first and the second robot, correspondingly. From (20.7) it is evident, that in this case the conflict between robots emerges only when the conditions $P^{[1]} = P^{[2]}$ and $\dfrac{i}{j} = \dfrac{t_{[2]}^0}{t_{[1]}^0}$ are valid.

We want to dwell on the relations determining fellowship (see Chapter 4) of robots with an absolute memory. In this case (under tantamount emotions) a number of education time steps necessary for achieving fellowship (concordance) between two sub-groups with equal fellowship values can be found by solving the following problem:

solve for

$$\min_{j \geq 1} (jq + R_0 - P_0), \qquad (20.8)$$

under $jq + R_0 - P_0 \geq 0$.

It is easy to see that this problem always has a solution, what means that robots with an absolute memory at any time can be brought to fellowship with any fellowship value preset.

Now let us solve the problem of developing the equivalent educational process (see Chapter 5) for robots with an absolute memory.

Obviously, in order to define the elementary education value $q$ corresponding to the equivalent process, we have to solve the following problem:

solve for:

$$\min_q J(q) = \min_q \sum_{j=2}^{n} \left[ R_j - R_1 - (j-1)q \right]^2. \qquad (20.9)$$

Problem (20.9) can be reduced to solving the equation $\frac{dJ(q)}{dq} = 0$ which expanded form is

$$\sum_{j=2}^{n} [R_j - R_1 - (j-1)q](j-1) = 0. \qquad (20.10)$$

It is easy to see that the solution $q$ of Eq. (20.10) is defined by

$$q = \frac{\sum_{j=2}^{n} R_j(j-1) - R_1 \sum_{j=2}^{n}(j-1)}{\sum_{j=2}^{n}(j-1)^2}.$$

## 21. ALGORITHM OF EMOTIONAL CONTACTS IN A GROUP OF ROBOTS

In this chapter we suggest a rule of mutual contacts between robots in their group.

In Chapter 2 we showed that the robot's education $R_i$ by the end of the $i$-th time step is specified by the formula

$$R_i = r_i + \theta_i R_{i-1}, \qquad (21.1)$$

where $\theta_i$ is the robot's memory coefficient which characterizes memorization of the education $R_{i-1}$ by the end of the $i$-th education time step.

Suppose robots contacting each other in a group randomly exchange emotions which initiate elementary educations.

Let $R_i^{[L]}$ be the education of the $L$-th robot by the end of the $i$-th time step, and also let $r_i^{[L]}$ be the elementary education corresponding to this time step. Similarly, let us introduce the corresponding educations $R_i^{[j]}$ and $r_i^{[j]}$ for the $j$-th robot.

Assume both robots are effected by the subject $S(t)$ initiating emotions $M_i^{[L]}$ (robot $L$) and $M_i^{[j]}$ (robot $j$).

Let us consider that if $R_{i-1}^{[L]} R_{i-1}^{[j]} < 0$ is valid then $M_i^{[L]} M_i^{[j]} < 0$ holds true, and the formula $R_{i-1}^{[L]} R_{i-1}^{[j]} > 0$ implies $M_i^{[L]} M_i^{[j]} > 0$.

The emotions $M_i^{[L]}$ and $M_i^{[j]}$ initiate the elementary education $r_i^{[L]}$ and $r_i^{[j]}$ correspondingly, and $r_i^{[L]} = \int_0^{t_i} M_i^{[L]}(\tau) d\tau$ and $r_i^{[j]} = \int_0^{t_i} M_i^{[j]}(\tau) d\tau$, where $t_i$ is the length of the *i*-th time step. Obviously the sign of the elementary education equals to the sign of the emotion generating this education, and vice versa.

Let us assume that the sign of the education by the end of the *i*-1$^{st}$ time step is equal to the current sign of the emotion during the *i*-th time step and the elementary individual education by the end of this time step.

Now let us introduce the following definition.

<u>Definition 21.1</u>. The suggestibility coefficient $k_i^{[j,L]}$ is the value permitting the emotion *i* of the robot *L* to be replaced by the corresponding emotion of the robot *j* multiplied by the value of this coefficient, if $\left|r_i^{[L]}\right| < k_i^{[j,L]}\left|r_i^{[j]}\right|$ with $k_i^{[j,L]} > 0$.

It is obvious that $k_i^{[j,j]} \equiv 1$.

Assume that when two robots come in contact and start communicating, the education of each of them (according to Formula (2.1)) satisfy the relations

$$R_i^{[L]} = \overline{r}_i^{[L]} + \theta_i R_{i-1}^{[L]}, \quad R_i^{[j]} = \overline{r}_i^{[j]} + \theta_i R_{i-1}^{[j]}, \text{ with:}$$

$$\overline{r}_i^{[L]} = \max\left\{\left|r_i^{[L]}\right|, k_i^{[j,L]}\left|r_i^{[j]}\right|\right\} sign \begin{cases} r_i^{[j]}, if \ k_i^{[j,L]}\left|r_i^{[L]}\right| = \max\left\{\left|r_i^{[L]}\right|, k_i^{[j,L]}\left|r_i^{[j]}\right|\right\} \\ r_i^{[L]}, if \ \left|r_i^{[L]}\right| = \max\left\{\left|r_i^{[L]}\right|, k_i^{[j,L]}\left|r_i^{[j]}\right|\right\} \end{cases},$$

$$\overline{r}_i^{[j]} = \max\left\{\left|r_i^{[j]}\right|, k_i^{[L,j]}\left|r_i^{[L]}\right|\right\} sign \begin{cases} r_i^{[L]}, if \ k_i^{[L,j]}\left|r_i^{[j]}\right| = \max\left\{\left|r_i^{[j]}\right|, k_i^{[L,j]}\left|r_i^{[L]}\right|\right\} \\ r_i^{[j]}, if \ \left|r_i^{[j]}\right| = \max\left\{\left|r_i^{[j]}\right|, k_i^{[L,j]}\left|r_i^{[L]}\right|\right\} \end{cases},$$

$k_i^{[j,L]}$ the suggestibility coefficient of *j*-th robot's emotions to the robot *L*,

$k_i^{[L,j]}$ the suggestibility coefficient of *L*-th robot's emotions to the robot *j*,

$k_i^{[j,L]} > 0$, $k_i^{[L,j]} > 0$.

Let us introduce the following definitions.

<u>Definition 21.2</u>. With $k_i^{[j,L]}\left|r_i^{[L]}\right| = \max\left\{\left|r_i^{[L]}\right|, k_i^{[j,L]}\left|r_i^{[j]}\right|\right\}$ satisfied the *j*-th robot is called the agitator.

<u>Definition 21.2</u>. Re-education (re-bringing) of a robot is a sign reversal of the robot's individual education.

Obviously, signs of individual educations of robots in a group can reverse only if there are both robots with oppositely signed educations and robots-agitators.

According to Theorem 3.1 proved in Chapter 3, a conflict in a group occurs only if the sum education of this group equals to zero. Based on this we worked out the following theorem.

Theorem 21.1. A conflict in a group of robots can occur only if initial educations of these robots are oppositely signed and if there are agitators in this group.

This opens a way to software modeling of an emotional behavior of a closed group of intercommunicating robots. The input parameters of the corresponding software for modeling are supposed to be memory coefficients of each of the robots in this group, their initial individual educations and paired suggestibility coefficients. As the software runs the emotions of robots are initiated at random and so occur the corresponding elementary educations due to random contacts of robots. As a result we may obtain the computed sum education specifying conflicts in the group, as well as individual educations of each robot in this group. Due to numerical experiments it is possible to find critical values of suggestibility coefficients and memory coefficients causing conflicts in the group of robots after several paired contacts (contacts between two robots).

An algorithm of a robots' behavior in a group with a leader differs from an algorithm of a robots' behavior in a group without a leader due to the fact that in the first case while selecting a robot-educator which is a major agitator in the group it is necessary to find the order number of the greatest value of robots' individual educations. A robot with this number is supposed to act the part of a permanent agitator-and-leader.

## 22. ON INFORMATION ASPECTS OF E-creatures

Currently U.S. researchers discuss the question concerning creation of an electronic copy of a human being which can be called an E-creature [1].

We tried to study this idea of our American colleagues in terms of information.

Let us make a series of remarks:

1. There is no human being with an absolute memory, i.e. he\she always forgets a part of perceived information as this is his\her natural feature.

2. A human being is able to accumulate information – without forgetting immediately a part of it – by finite portions.

Now let us give the following definitions:

Definition 22.1. A portion is an amount of new information which is remembered completely by a human being.

Definition 22.2. An information time step is an arrival time of a portion.

Let us note one obvious property of the portion: a number of bits $s_i$ in the portion $i$ is limited, i.e. there is such $q$ for which the inequalities

$$s_i \leq q, \quad q \geq 0, \quad i = \overline{0, \infty}$$

always hold true.

Let us record the following formula according to the methods given in Chapter 2:

$$S_{i+1} = s_{i+1} + \lambda_{i+1} S_i, \qquad (22.1)$$

with: $i$ the number of the information time step, $i = \overline{0, n}$; $s_{i+1}$ the $i+1^{st}$ portion, $S_{i+1}$ the total amount of information memorized by a human through $i+1$ information time steps, $\lambda_{i+1}$ the human information memory coefficient (characterizes the part of total memorized information which was received during the $i$ previous information time steps).

Obviously the human information memory coefficient corresponding to the end of the information time step satisfies the relation $\lambda_i \in [0, 1)$ where there is $\lambda$ with

$$\lambda \geq \lambda_i, \quad i = \overline{0, \infty}, \quad \lambda \in (0, 1).$$

By virtue of the information property, $s_i \geq 0$ holds true, consequently all the accumulated information is greater than or equal to zero.

Suppose we have an electronic copy of a human created. Let us prove one of the information properties of this copy.

Theorem 22.1. The total information $S$ which can be memorized by the processor of the human-like copy is limited.

Proof. Applying the methods given in Chapter 2, portion properties and Eq. (22.1) we easily obtain the inequality

$$S_{i+1} \leq q \frac{1 - \lambda^{i+1}}{1 - \lambda}. \qquad (22.2)$$

Proceeding to the limit in Ineq. (22.2) with an infinite increase of time steps (time of existence of an immortal human) we get the chain of relations

$$S = \lim_{i \to \infty} S_i \leq q \lim_{i \to \infty} \frac{1 - \lambda^i}{1 - \lambda} = \frac{q}{1 - \lambda} < \infty.$$

Thus, the theorem is proved.

Corollary 22.1. It is impossible to create an E-creature with a nonabsolute memory which would be able to accumulate information infinitely.

Its proof is evident from the formulation of Theorem 22.1.

So, we can conclude that it is *impossible* to create the *only infinitely existing E-creature which would be an evolving copy of a human being* (*at least, in terms of information*).

An immortal (infinite) electronic creature able to accumulate information infinitely [1] is possible only in case if it has an *absolute* information storage (information memory) with the conditions $\lambda_i \equiv 1, \quad i = \overline{1, \infty}$ satisfied; but this sort of creature would have nothing to do with a human being, forgetful and oblivious; this sort of creature could be called just a robot unit.

For the infinite information evolution of the E-creature with an absolute memory we can state that it is necessary that the information from a chip of the "ancestor" E-creature with the nonabsolute memory shoul be downloaded to a chip of the "successor" E-creature (also with the nonabsolute memory) when the amount of the accumulated information becomes close to $S$. For the purpose of further data accumulation by the E-creature (which is a copy of a human being with a nonabsolute memory) it is necessary to re-download all the information from the ancestor's chip to the chip of the successor on a regular basis, i.e. $s_0$ is supposed to be equal to $S_k$ with $k$ the number of information time steps performed by the E-creature in the full course of its existence.

Let us note one property of memory information coefficients varying during the information time step length $t$ with $t \in [t_i, t_{i+1}]$.

Theorem 22.2. $\lambda_{i+1}(0) = 1$.
Proof. Similarly to (22.1), let us write the formula

$$S_{i+1}(0) = s_{i+1}(0) + \lambda_{i+1}(0)S_i. \tag{22.3}$$

But at the initial moment of the information time step the relations
$$S_{i+1}(0) = S_i, \quad s_{i+1}(0) = 0 \tag{22.4}$$
hold true.
Substituting (22.4) into Relation (22.3) and solving the obtained equation relative to $\lambda_{i+1}(0)$ we get $\lambda_{i+1}(0) = 1$, which was to be proved.

Let us define a linear dependence allowing approximately describe the change in the memory information coefficient during the information time step.
Obviously, $S_{i+1} = s_{i+1} + \lambda_{i+1}(t_{i+1})S_i$. Consequently,

$$\lambda_{i+1} = \lambda_{i+1}(t_{i+1}) = \frac{S_{i+1} - s_{i+1}}{S_i}. \tag{22.5}$$

holds true.
Suppose that $\lambda_{i+1}(t) = at + b$ holds true.

By Theorem 22.2 and Formula (22.5) the system of linear equations
$$\lambda_{i+1}(0) = b = 1, \tag{22.6}$$

$$\lambda_{i+1}(t_{i+1}) = \lambda_{i+1} = a(t_{i+1} - t_i) + b. \qquad (22.7)$$

holds true.
Solving this system of equations (6) – (7) we get

$$a = \frac{\lambda_{i+1} - 1}{t_{i+1} - t_i}, \quad b = 1.$$

Thus we can write down the following formula

$$\lambda_{i+1}(t) \approx \frac{\frac{S_{i+1} - s_{i+1}}{S_i} - 1}{t_{i+1} - t_i} t + 1,$$

with $t \in [t_i, t_{i+1}]$.

It is easy to see that many proposition and provisions of the emotional robot theory given in the previous chapters can be easily adapted to the aspects of data accumulation by the E-creature. We suggest that you, our dear reader, should do it yourself as some brain exercises for pleasure at your leisure.

## 23. SOFTWARE REALIZATION OF SIMPLE EMOTIONAL ROBOT's BEHAVIOR

In order to illustrate the theory given in Chapter 2 let us set the task of developing software which would model the emotional behavior of a robot taking and responding audible cues (sounds) which are put in this software through a microphone plugged to a computer. Assume this computer program is to execute the following: according to a sound amplitude the program determines a type of "smile" which is outputted by a computer monitor as a response (reaction) to the sound effect (so finally we will see different "shades" of sad or happy smiles).

### 23.1. INPUT PARAMETERS OF SOFTWARE

Assume the modeled robot is uniformly forgetful. As the input parameters for the model implemented by this software we use the robot's memory coefficient $\theta$ equal to some constant value from 0 to 1, and the time step.

### 23.2. ALGORITHM FOR MODELLING ROBOT's MIMIC EMOTIONAL REACTION

In this section we suggest an algorithm which helps to model the mimic emotional reaction of the robot effected by a sound (audio signal).
This algorithm represents a sequence of steps which would make a robot (software) emotionally react (mimic) to sounds produced by a human, animal, etc.
Let us present this algorithm as the following sequence of steps with some explanations:

1. Convert analog sound signals received from a microphone, to a sequence of numbers representing momentary values of a signal amplitude. Analog-digital [A/D] converters are pretty suitable devices for such a purpose. And the conversion method itself is called the pulse-code modulation.

2. Collect data necessary for the following analysis.

3. Analyze and aggregate the collected data.

4. Reveal and evaluate the degree of the predefined emotional stimuli. In other words, specify subject values effecting the robot (software). Predefined sound characteristics can be used as the emotional stimuli; sound characteristic data collection is to be done at Step 2 and 3.

5. Compute momentary emotional characteristics of the robot (software) on the basis of the emotion and education model considered in Chapter 2.

6. Compute elementary educations on the basis of momentary emotional characteristics.

7. Compute the education on the basis of elementary educations and the robot's (software's) memory coefficient which is to be preset before the algorithm is started.

8. Enjoy a visualization of the robot's (software's) emotions based on the computed education.

Let us consider each step of the algorithm in more detail.

Step 1. In order to go through the 1-st step we need an analog-digital converter. Every modern soundcard is usually equipped with it, so in order to get an access to it we need to interact with a soundcard driver. It can be fulfilled in a variety of ways, some of which we will consider below. Generally speaking, it is very important to setup the conversion itself, i.e. its characteristics. It is necessary to select and preset the sound sampling frequency, signal discreteness, number of channels and other characteristics.

Step 2. This step deals with data collection from the soundcard in the course of the pulse-code modulation. The data can be stored in a variety of ways, e.g. in files of different formats, or just store the internal data structure. However here we should take into consideration that the data size (even if the interaction of the stimulant and the robot (software) is very brief) may grow pretty big. E.g. with the sampling frequency of 22050 Hz, discreteness of 8 bits, mono channel and 10-second stimulant – robot interaction, the robot (software) is supposed to receive 220500 bytes from the soundcard.

Step 3. The data is analyzed and aggregated, i.e. some certain preset characteristics are computed on the basis of a whole data bulk or just a part of it.

Step 4. The 4-th step is matching, which means that on the basis of certain values of the computed characteristics evaluation of subjects' values takes place. Correct matching is achieved experimentally.

Steps 5 is similar to 4, only at this step the momentary emotional characteristics are matched to the degrees of the effecting subjects. Correct matching is achieved experimentally as well.

Steps 6 and 7 imply computations based on the mathematical model formulas described in Chapter 3.

At the final step of the algorithm the robot's emotion is to be expressed visually. This can be fulfilled by some of the ways of emotion visualization (e.g. a 'smile').

Also we should note the following. If we want to develop an 'interactive' robot (software) i.e. the robot responding to sounds instantly then data collection and data processing are to be executed simultaneously.

Thus the 2-nd step of the algorithm is to be executed simultaneously to Steps 3 – 8.

### 23.3. SoundBot SOFTWARE ARCHITECTURE

Let us examine an architecture of the developed software SoundBot [11] implementing the algorithm given above (Fig. 23.1.). Figures in circles mean the steps of the algorithm.

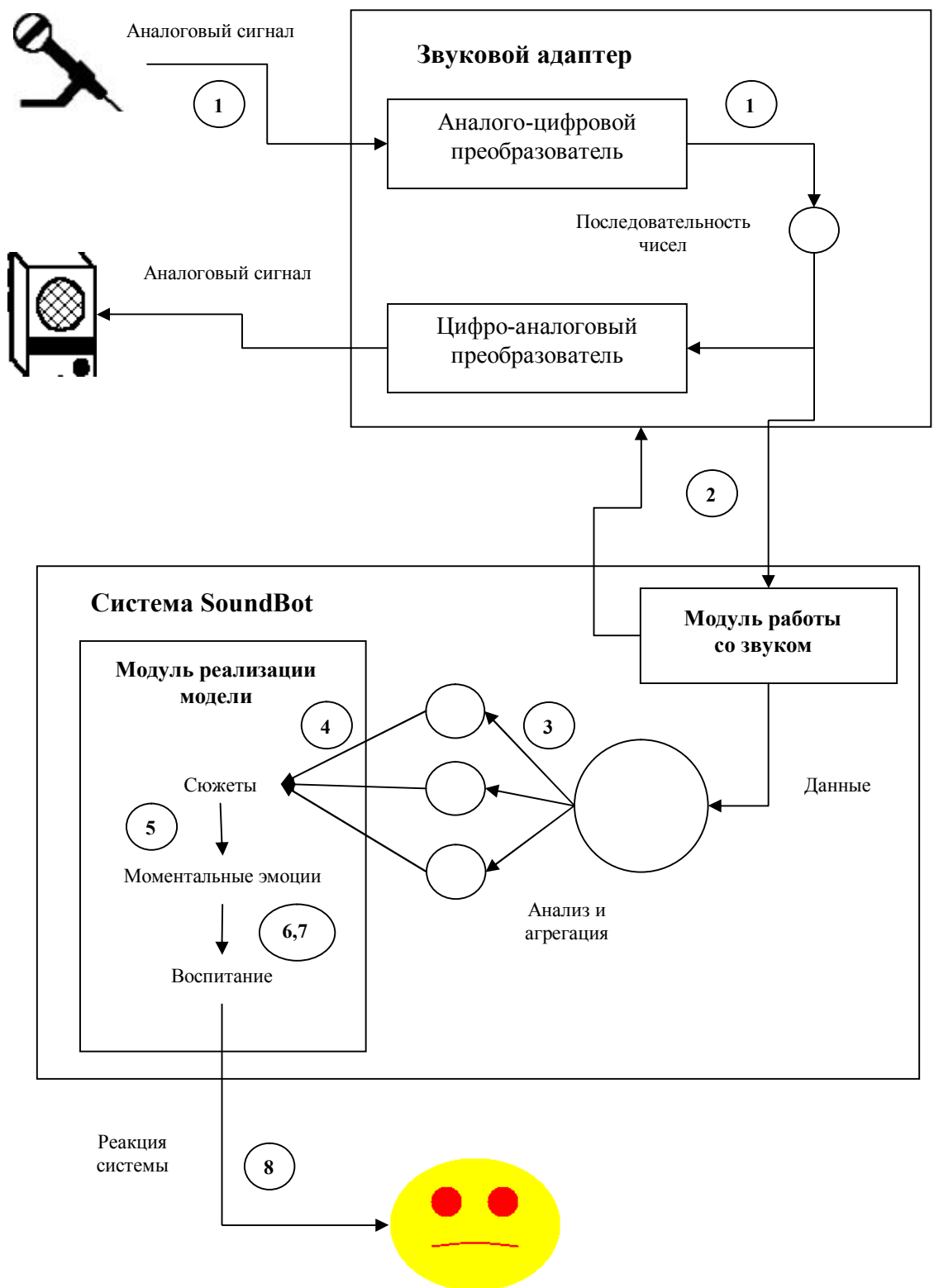

Fig. 23.1. Architecture of SoundBot software

It is easy to see that the architecture is directly associated with the algorithm given above. It includes two modules:

1. A sound module, which is responsible for interaction with the soundcard and collection of the necessary numerical data.

2. An implementation module, which is responsible for implementation of the given mathematical model of emotions and education, it also computes the smile parameters to show the mimic emotional response of the system.

The data is processed, analyzed and aggregated directly between the modules. Both modules function simultaneously for the system to be interactive.

Now let us examine main features, operation principles and visual interface of this software.

### 23.4. MAIN FEATURES OF SoundBot

This software is written in C++ using Visual Studio 2008 development environment. It works on IBM PC compatible computers under Windows XP and elder OS. The software also requires .NET Framework 2.0. The exe file size is 100 Kbytes.

The major functions of this software are the following:
1. SoundBot is able to detect main capabilities of PC multimedia devices.
2. SoundBot is able to play wav files.
3. SoundBot is able to record sounds in wav files (mono only).
4. SoundBot can perform an emotional response to the played wav files.

SoundBot can emotionally response in an interactive mode to the sounds inputted via a microphone.

### 23.5. SoundBot OPERATION PRINCIPLES

Major operation principles of SoundBot which are to be viewed in details are:
1. Sound module operation,
2. Principle of simultaneous operation of both modules.
3. Emotional stimuli considered by the software and principles of their degree assignment.

As it was said before, there are a variety of ways for working with a soundcard. The methods considered above use system libraries of MS Windows, so these methods can be used only with this OS.

The simplest approach is to use MCI command-string interface or MCI command-messages interface. MCI is a universal interface independent of hardware characteristics. MCI is meant for controlling multimedia devices (soundcards and videocards, CD- and DVD-ROMs) [12, 13].

In most cases capabilities of this interface meet the needs of any multimedia applications used for recording and playing audio or video files. But it has a drawback: the data received from the soundcard cannot be read and processed interactively. It means that this method will not work here.

This approach is based on the MCI command-string interface or MCI command-message interface and the drawback of this method can be overcome if we use a low level interface.

The low level interface can be used for playing wav files as follows. First, the wav file header is read and its format is checked, the output device is opened and the sound data format is specified. Next, the audio data blocks are read directly from the wav file to get prepared by a special function for output and then they are passed to the driver of the output device. The driver puts them out to the soundcard [12, 13]. The application totally controls the playback process because it prepares the data blocks in RAM itself.

The audio data is recorded the same way. First, the input device is to be opened and the audio file format is to be specified to the device. Next, one or more blocks of RAM are to be reserved and the special function is to be called. After that, as the need arises, the prepared blocks are passed to the input device driver which fills them with the recorded audio data [12, 13].

For the recorded data to be saved as a wav file the application has to generate and record the file header and audio data to the file from the RAM blocks prepared and filled by the input device driver.

The low level interface requires all the record-and-playback details to be very thoroughly considered, as opposed to the MCI interface where most of parameters are just taken by default. These extra efforts are compensated with pretty good flexibility and the opportunity to work with the audio data in real time [12, 13].

To provide the interactive mode of the SoundBot, i.e. make it interact with a user in real time, its modules have to operate simultaneously.

Each SoundBot's module is executed as a separate thread and it makes possible the following:

1. The software can simultaneously receive new data from the soundcard and analyze it for further computing of the education which reflects the emotional state.

2. The software can simultaneously play, record and select the audio data for its analysis.

Besides, the visualization of mimic emotional response is also executed as a separate thread to make it drawn as fast as possible.

Still the SoundBot considers only one emotional stimulus (subject) which is amplitude of the effecting audio signal. Every audio signal count generates stimulations in the SoundBot system and initiates momentary emotions according to the sine-shaped emotion function. Subjects are matched to emotions by value ranges specifying what subject initiates positive emotions and what subject initiates negative ones.

## 23.6. SoundBot VISUAL INTERFACE

A main window includes two inlays: the first one deals with playback and training of the SoundBot system on .wav samples (Fig. 23.2).

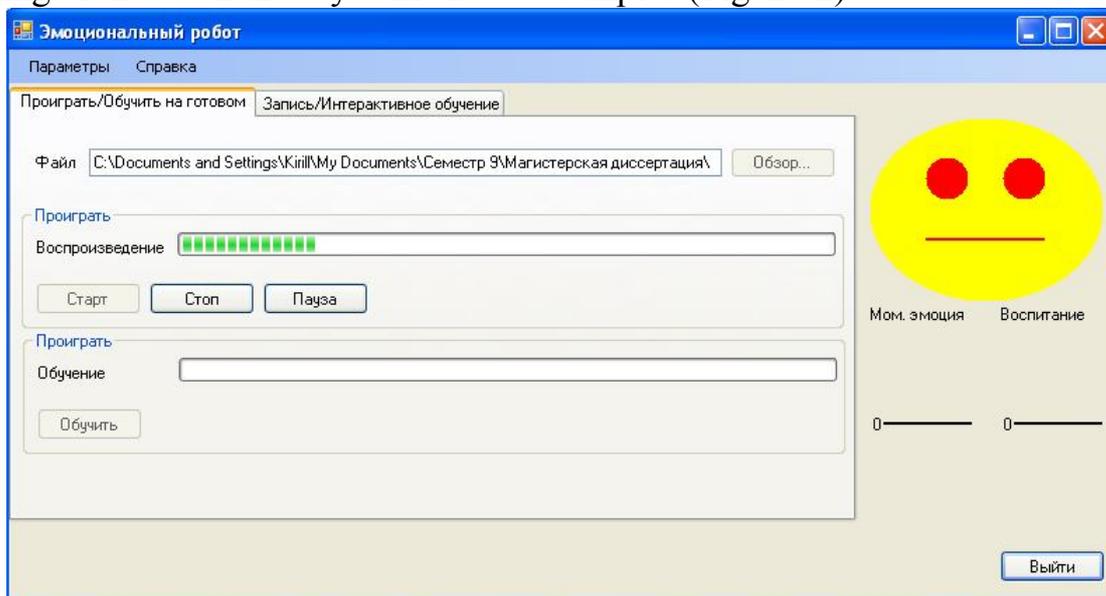

Fig. 23.2. First inlay in the main window of the SoundBot software.

The second inlay is used for recording wav files and interactive communication with the SoundBot system (Fig. 23.3).

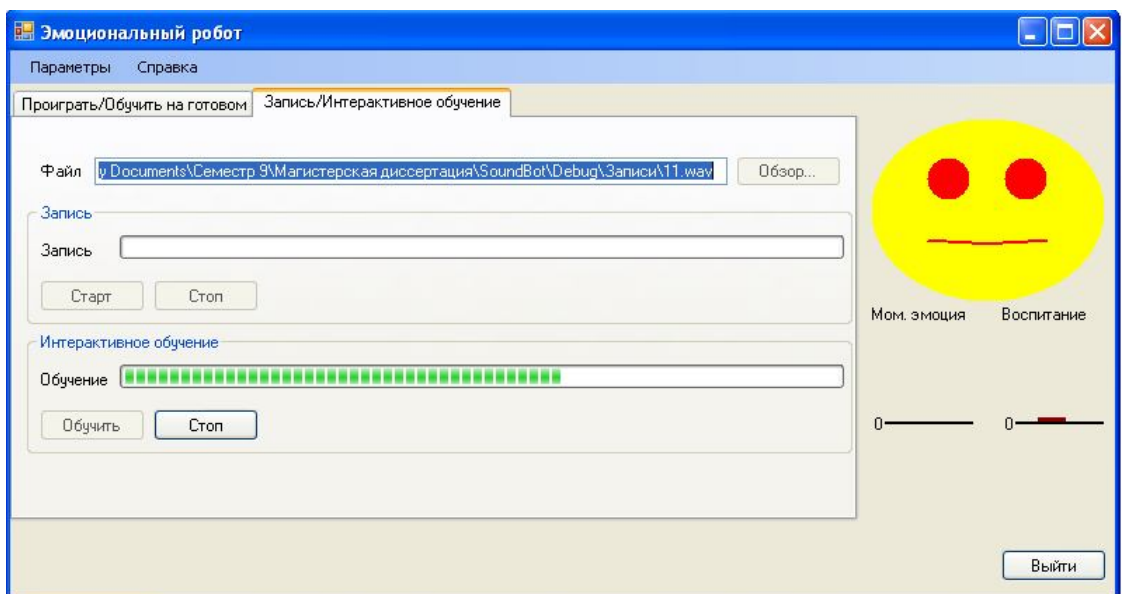

Fig. 23.3. Second inlay in the basic window of the SoundBot software.

Besides, the main window shows a smile expressing the emotional response of the modeled robot and the current value of the momentary emotion and education.

In a main menu we may set the major parameters (parameters of the emotion math model, parameters of operation principles and parameters of audio data processing).

Below we show a couple of dialogue windows for setting up different parameters (Fig.23.4 and 23.5).

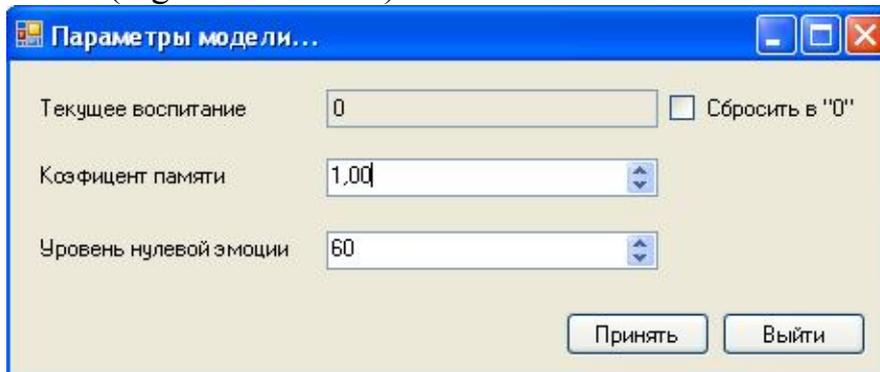
Fig.23.4. Model parameters

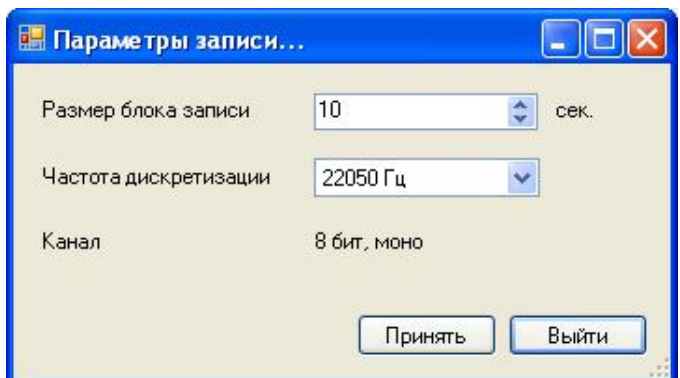
Fig.23.5. Record parameters

To find out characteristics of pulse-code conversion supported by the soundcard we are to select the option "Info" –> "Driver parameters…" of the main menu (this is strongly recommended for the correct record parameters settings especially when the software is run for the first time).

After you submit the settings you will see a window containing the description of multimedia hardware (Fig.23.6).

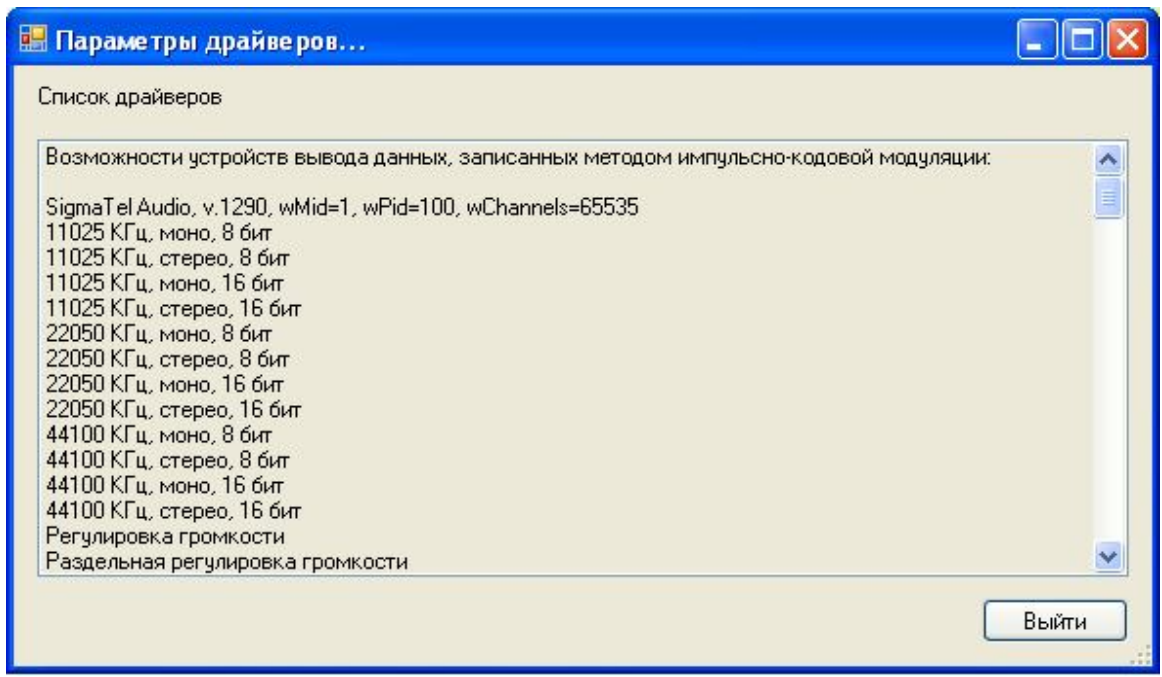

Fig.23.6. Multimedia hardware Parameters

The suggested algorithm can be used for building emotional robots. But the input audio data should be analyzed more thoroughly to single out as much stimuli as possible. That is why the SoundBot system can be considered as the first approximation of emotional robot software.

Also it should be considered that both the algorithm and Soundbot itself are meant for interaction with only one user. Interaction with several users requires some other much more complicated mathematical model.

The described software can be applied, for instance, for proper communication and rehabilitation of hearing-impaired patients, or used by actors for placing a voice outside an opera house. This software can also be used for predicting the emotional reaction of other people to the user's behavior (the software response shows the possible reaction of the surrounding people).

# CONCLUSION

We hope you managed to read this book through. The authors made an attempt to build up and describe the virtual reality of emotional robots.

Concerning real mental processes of living organisms, it is not easy to define dependencies between emotions and time, and, perhaps, in the general case, this problem is unsolvable. But in the process of building robots a roboticist can preset mathematical functions of emotions altering with time (same as memory coefficients, and derivatives of emotion functions). In this case the theory given in this book allows of designing robots with the preset psychological characteristics, with further analyzing and computing of emotional behavior of robots on the basis of numeric data read in their memory.

As an example, below we give a description of a closed chaotic virtual reality of emotional robots based on software implementation of mathematical models shown in this book. In this description we use the terms defined above.

*Let the virtual reality include some finite number of robots. Each of robots has its own memory with its special individual memory coefficients. In their virtual reality robots effect upon each other with different subjects in a random way to initiate emotions and alter each other's educations. Robot the educator (the one from which emotions are passed to the educatee) is that with the greatest education modulo. Concordance groups – 'fellowships' of robots occur as a result of emotional contacts between robots, the greater their fellowship value, the more united is the group. Some groups may get into conflicts with each other. These conflicts emerge when sum educations of the groups became equal to zero. Each robot has a goal which is common for their reality in a whole. As a result of this goal presence in the course of time the leaders may appear which are robots with the greatest willpower and best abilities. Education effectiveness of each robot is characterized by the education process efficiency coefficient. Finding their efficiency coefficients can help us to select robots with natural characteristics making them the most educationally inclined. Some of robots feature satiated education; when these robots get to some certain level of satiety, emotional effect of other robots upon them stops. If there are robots which do not have education satiety in this virtual reality, then other robots educate them in the most active way, and this causes leaders to occur in the robots' community. Based on equivalent processes developed for each of robots with further ranking of limit educations, a leader of the robots' community defines its distant successor to be a new leader in future. The robots may get ill due to some software faults or computer virus attacks. A physician in this robots' community heals its ill inhabitants by correcting their emotions. As robots-members of this community keep communicating and interacting with each other their educations alter with the course of time. This causes the leaders to change and new fellowships and conflicting groups to occur. This is the way emotional robots live in their virtual reality.*

This book appeared as a result of investigations described in [3, 11 – 33], it includes new results and prepares a basis for new problems.

We hope that this book is useful for roboticists and program developers designing software for emotional robots and their groups.

Any your ideas and opinions about this book are welcome. Please feel free to e-mail to the authors at ogpensky@mail.ru or  kirillperm@yandex.ru .